\PassOptionsToPackage{unicode}{hyperref}
\PassOptionsToPackage{hyphens}{url}
\documentclass[
  11pt,
]{article}
\usepackage{amsmath,amssymb}
\usepackage{iftex}
\ifPDFTeX
  \usepackage[T1]{fontenc}
  \usepackage[utf8]{inputenc}
  \usepackage{textcomp} % provide euro and other symbols
\else % if luatex or xetex
  \usepackage{unicode-math} % this also loads fontspec
  \defaultfontfeatures{Scale=MatchLowercase}
  \defaultfontfeatures[\rmfamily]{Ligatures=TeX,Scale=1}
\fi
\usepackage{lmodern}
\ifPDFTeX\else
\fi
\IfFileExists{upquote.sty}{\usepackage{upquote}}{}
\IfFileExists{microtype.sty}{% use microtype if available
  \usepackage[]{microtype}
  \UseMicrotypeSet[protrusion]{basicmath} % disable protrusion for tt fonts
}{}
\makeatletter
\@ifundefined{KOMAClassName}{% if non-KOMA class
  \IfFileExists{parskip.sty}{%
    \usepackage{parskip}
  }{% else
    \setlength{\parindent}{0pt}
    \setlength{\parskip}{6pt plus 2pt minus 1pt}}
}{% if KOMA class
  \KOMAoptions{parskip=half}}
\makeatother
\usepackage{xcolor}
\usepackage[margin=1in]{geometry}
\usepackage{color}
\usepackage{fancyvrb}

\DefineVerbatimEnvironment{Highlighting}{Verbatim}{commandchars=\\\{\}}
\newenvironment{Shaded}{}{}

\newcommand{\AttributeTok}[1]{\textcolor[rgb]{0.49,0.56,0.16}{#1}}

\newcommand{\ExtensionTok}[1]{#1}

\newcommand{\NormalTok}[1]{#1}

\usepackage{longtable,booktabs,array}
\usepackage{calc} % for calculating minipage widths
\usepackage{etoolbox}
\makeatletter
\patchcmd\longtable{\par}{\if@noskipsec\mbox{}\fi\par}{}{}
\makeatother
\IfFileExists{footnotehyper.sty}{\usepackage{footnotehyper}}{\usepackage{footnote}}
\makesavenoteenv{longtable}
\usepackage{graphicx}
\makeatletter
\newsavebox\pandoc@box
\newcommand*\pandocbounded[1]{% scales image to fit in text height/width
  \sbox\pandoc@box{#1}%
  \Gscale@div\@tempa{\textheight}{\dimexpr\ht\pandoc@box+\dp\pandoc@box\relax}%
  \Gscale@div\@tempb{\linewidth}{\wd\pandoc@box}%
  \ifdim\@tempb\p@<\@tempa\p@\let\@tempa\@tempb\fi% select the smaller of both
  \ifdim\@tempa\p@<\p@\scalebox{\@tempa}{\usebox\pandoc@box}%
  \else\usebox{\pandoc@box}%
  \fi%
}
\def\fps@figure{htbp}
\makeatother
\providecommand{\tightlist}{%
  \setlength{\itemsep}{0pt}\setlength{\parskip}{0pt}}
\usepackage{fvextra}
\DefineVerbatimEnvironment{Highlighting}{Verbatim}{commandchars=\\\{\},breaklines,breakanywhere,fontsize=\small}
\fvset{breaklines,breakanywhere}
\usepackage{needspace}
\usepackage{xurl}
\usepackage{etoolbox}
\AtBeginEnvironment{longtable}{\raggedright\footnotesize\setlength{\tabcolsep}{3pt}}
\usepackage{caption}
\ifdefined\DeclareUnicodeCharacter\DeclareUnicodeCharacter{2212}{\ensuremath{-}}\DeclareUnicodeCharacter{00B7}{\ensuremath{\cdot}}\fi
\usepackage{bookmark}
\IfFileExists{xurl.sty}{\usepackage{xurl}}{} % add URL line breaks if available
\hypersetup{
  pdftitle={kw-bench: Breaking Frontier Harnesses on Real-World Tasks},
  pdfauthor={George Salapa, G.S. s.r.o., PwC Austria},
  hidelinks,
  pdfcreator={LaTeX via pandoc}}

\title{kw-bench: Breaking Frontier Harnesses on Real-World Tasks}
\usepackage{etoolbox}
\makeatletter
\providecommand{\subtitle}[1]{% add subtitle to \maketitle
  \apptocmd{\@title}{\par {\large #1 \par}}{}{}
}
\makeatother
\subtitle{Harness Paradigm for Knowledge Work}
\author{George Salapa, G.S. s.r.o., PwC Austria}
\date{}

\begin{document}
\maketitle

\section*{Abstract}\label{abstract}
\addcontentsline{toc}{section}{Abstract}

Each new generation of frontier model pushes the edge of what is
possible. Inside a harness --- a loop, a shell, a filesystem, the
ability to write and run its own code --- a model can take on
increasingly difficult work. Software engineering was the first ground
it took. It will not be the last.

A growing body of work treats the coding-agent harness as infrastructure
for knowledge work rather than a coding tool, and a new wave of
benchmarks tests it there. Most of them are generative: build a
financial model, a memo, a spreadsheet from a brief, judged by a rubric
or by another model. Few test the expensive part of knowledge work,
which is investigative. In audit, tax, due diligence, regulatory
reporting, claims or any dispute over numbers, the answer already exists
--- hidden in incomplete and inconsistent evidence spread across systems
--- and has to be recovered and defended to the cent.

We introduce kw-bench, a benchmark built from real engagements
specialists described to us. Its first task family is a five-year VAT
reconstruction across three generations of SAP --- legacy ECC, a partial
S/4HANA migration and a consultant's redesign --- delivered as raw
system extracts: page headers, local number formats, partial link
tables, no documentation of how anything connects. It asks a harness to
do what a team of specialists does over weeks: reverse-engineer how
three systems recorded the same business, invent its own method to
parse, join and reconcile them, keep dozens of weak signals in view for
hours, avoid the traps that invent exposure, and defend every number to
the cent --- then hand over the full package on company templates and a
reusable tool tested on data it has never seen. Grading is
deterministic, with no LLM judge. It runs on Harbor.

We ran four harnesses --- Claude Code, Codex, OpenCode and micro-cc ---
on the same model, DeepSeek v4.1 Flash, headless, for two hours. The two
independent, model-agnostic harnesses delivered (OpenCode 0.542,
micro-cc 0.528); the two lab harnesses, away from their home model,
delivered nothing (0.00 each). No single step defeated any of them. What
defeated them was inertia: keeping many weak signals alive across a long
session. On a frontier model the task collapses. micro-cc and Claude
Code on Claude Opus 5.5 scored 0.9986 and 1.000, in 22 and 25 minutes,
for \$5.80 and \$5.89, both finding the exposure exact to the cent and
delivering the full package. A weeks-long engagement now costs minutes
and a few dollars; below the frontier, the harness decides how far a
model gets.

\begin{figure}
\centering
\pandocbounded{\includegraphics[keepaspectratio]{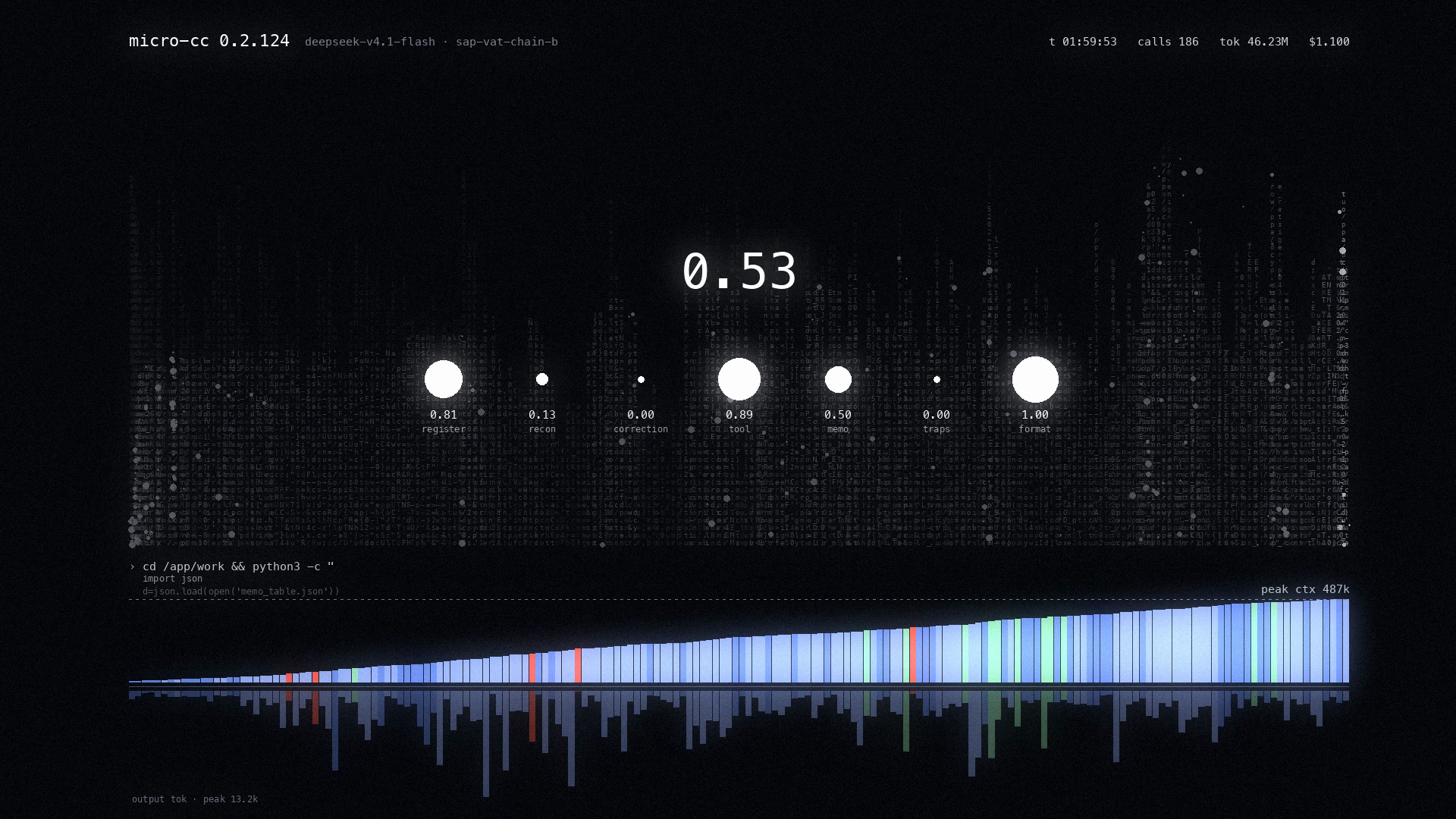}}
\caption{Figure 1. One run, every tool call: micro-cc 0.2.124 on
DeepSeek v4.1 Flash, two hours, 186 model calls. Each column band is one
tool call, drawn from its own command text. The bars at the bottom are
per-turn context (up) and output tokens (down); red marks a tool error,
green a deliverable written. The end card shows the seven component
scores and the total. The cost in the header is a list-price estimate.
The full run is in the supplementary video.}
\end{figure}

\section*{1. Introduction}\label{introduction}
\addcontentsline{toc}{section}{1. Introduction}

Frontier models have, by most accounts, collapsed the complexity of
software development. A less discussed fact is that the same ability ---
iterating on a task by writing and running code and storing the
artifacts of its work on a filesystem --- extends to anything with a
digital footprint. The agent harness was built for software engineering,
but the pattern is general: it turns a model from something that answers
into something that can do deep work. It can open a 60 MB extract, write
a parser for it, test a hypothesis, find it wrong and try again, without
the dataset ever sitting in its context. Increasingly, knowledge work is
a coding problem.

A growing body of work treats the harness as infrastructure rather than
a coding tool, and it converges on three findings. A
terminal-and-filesystem agent matches or beats more elaborate agent
architectures on enterprise work (arXiv:2604.00073), and handles simple
ERP tasks with no ERP-specific tooling at all, failing as complexity
rises through ignored constraints and overconfidence (arXiv:2604.13107).
On long-horizon tasks, the harness often affects performance more than
the model does (arXiv:2605.23950). And the harness --- code as the
operational substrate of the agent --- is becoming the unit an
organisation engineers, governs and distributes (arXiv:2605.18747,
arXiv:2607.25890; our own earlier paper arXiv:2608.20622). What this
work does not test is whether a harness can reach a defensible
conclusion from evidence. Its enterprise tasks are mostly operational
--- do X in system Y --- short, and checked by end state.

Outside the developer community this is still poorly understood. We
would cautiously say that knowledge workers, out in the wild, automate
the wrong thing: a single pass at the output layer, building the deck or
drafting the email. These tasks are tedious, undeniably, and a model
should take them: nobody needs to spend an evening aligning boxes on a
slide. But the message on each slide has to be the specialist's own
understanding, in the specialist's own voice. You can outsource the
making; you cannot outsource the understanding. Handing over the deck
saves an hour. Handing over the message costs the voice.

The tasks worth pointing a model at are somewhere else entirely.

The cognitively taxing part of knowledge work sits several layers
underneath the output. It is pulling information out of deep systems and
synthesising it across many hops --- work that used to mean days in
Excel, in SAP transactions, in Snowflake or Databricks. It is the
auditor tracing a revenue line back through three ledgers, the
due-diligence team rebuilding a target's working capital from raw
postings, the claims analyst matching policies to payments across two
migrated systems, the controller explaining a budget overrun from
purchase orders, goods receipts and payment runs. Take one case in
detail: filed tax returns that have to be verified across several
accounting systems, over several years, against emails, ERP data, local
workbooks and an adviser's memos. Proving what went wrong takes a team
of specialists weeks: pulling extracts, writing joins nobody documented,
chasing a change through change logs and posting runs, reconciling to
the cent, and writing it up so that it survives an audit. That work is
slow, repetitive in method and unique in every case. It is exactly the
shape of work a harness was built for.

We use one word for the person who owns that work: the
\emph{specialist}. We mean it loosely --- anyone who knows, or has the
potential to know, a problem intricately, rather than stopping at its
surface layer.

The knowledge-work benchmarks that exist test something else. They are
predominantly generative: produce a financial model, a memo, a slide
deck or a spreadsheet from a brief, judged by a rubric or by another
model (Section 2). We see three reasons this misses the expensive part
of the work.

First, the answer is not hidden. In a generative task the brief contains
what is needed, and quality is judged on form and plausibility. In real
analysis the answer exists, in the data, and the work is recovering it.

Second, there is no anchor. Rubric and judge grading rewards a
convincing story. One of the cases specialists brought to us was exactly
this: a full booking dump and a question, with nothing to test any
answer against --- ``the first suggestion any model makes is as valid as
the fifteenth''. A benchmark built like that measures persuasion, not
analysis.

Third, the inputs are clean: single tables, consistent formats,
documented keys. The hard part of real work is the mess --- exports with
page headers, local number formats, three generations of one system.

Real cases are investigative and reconstructive. Forensics is only our
specimen; the same shape runs through audit, tax, due diligence,
regulatory reporting and every dispute over numbers. The answer exists,
hidden in incomplete, inconsistent evidence spread across systems, and
has to be defended to the cent. The difficulty is not volume. It is
holding a long chain of reasoning accountable to a large and messy body
of evidence: every claim traced to a table, every number reproducible,
and no data point can, of course, be invented. A model cannot keep five
years of postings `in its head'; through the harness it moves its memory
out of the context window into code, intermediate files and checks it
can rerun. The harness is what keeps the model honest against the
evidence.

This is what we built kw-bench to test. It is not a puzzle set: each
task mirrors one real engagement end to end, from the email that
triggers it to the package a specialist would sign (Section 3). The
trigger is a Head of Tax's email with a deadline. The data arrives the
way IT delivers it. The system history is the real history --- legacy
ECC, a migration to S/4HANA with a partial number map, a consultant's
redesign that links documents through a custom field and hand-typed
references. The rules are stated the way an adviser states them, in a
position memo, so the test is reconstruction and application rather than
recall of obscure law. The anchor is the real anchor: the returns that
were actually filed. And the output is the real output --- registers and
reconciliations in the company's templates, a correction upload in the
compliance tool's format, a two-page memo on letterhead, and a tool the
team can rerun every month, tested on data it has never seen.

There is a cost argument as well. Some organisations now spend very
large token budgets, at a scale that would employ many people, on work
of relatively low margin. Our results point the other way. A harness and
a model give a specialist a near-unbounded ability to move across a
complex problem --- to navigate a model through building the exact
one-off system that solves it: check the direction, nudge, discard,
rebuild, at a cost that used to be prohibitive. In our runs the weeks of
evidence work behind one tax decision cost between a quarter of a dollar
and six dollars per harness-model pair. Dollars and cents, not millions.

This paper makes four contributions.

\begin{enumerate}
\def\labelenumi{\arabic{enumi}.}
\tightlist
\item
  \textbf{kw-bench}, to our knowledge the first benchmark that combines
  an SAP-style schema across three system generations, exported as real
  extracts arrive; a hidden answer checkable to the cent; traps that
  penalise invented findings; a full professional deliverable package;
  and reusable code graded on hidden data. Grading is deterministic,
  there is no LLM judge, and it runs on Harbor (Section 3).
\item
  \textbf{Deep knowledge work is collapsible by a frontier harness}, not
  only generative work. A forensic VAT engagement that takes a team
  weeks collapses into about 25 minutes and six dollars, exact to the
  cent, with every deliverable a specialist would sign (Section 5.4).
\item
  \textbf{Below the frontier, the harness is the decisive factor.} Same
  model, same task, same two hours: the two independent harnesses
  delivered about half the score, the two lab harnesses nothing. We
  trace the gap to concrete harness engineering --- what goes back into
  the context, tools and defaults that assume the home model, recovery
  from a broken turn, verification before declaring done (Sections 4 and
  5.3).
\item
  \textbf{The failure mode is inertia.} No single step defeated any
  harness; what defeated them was keeping many weak signals alive across
  a long session. This is exactly where the specialist's role moves:
  from doing the evidence work to holding the checklist and steering
  (Section 4.4).
\end{enumerate}

\section*{2. Background}\label{background}
\addcontentsline{toc}{section}{2. Background}

\textbf{The harness as infrastructure.} Terminal Agents Suffice for
Enterprise Automation (arXiv:2604.00073) shows that an agent with a
terminal and a filesystem matches or outperforms MCP- and GUI-based
architectures on enterprise tasks, and argues that simple programmatic
interfaces plus strong foundation models could be the backbone of
enterprise automation. Can Coding Agents Be General Agents?
(arXiv:2604.13107) puts a coding agent in front of an open-core ERP
system: simple tasks succeed with no ERP-specific tooling, and failures
grow with complexity --- lazy heuristics, hallucination at the business
layer, ignored constraints, overconfidence. Its ERP is clean, single and
well documented; real analysis starts where the documentation ends. Stop
Comparing LLM Agents Without Disclosing the Harness (arXiv:2605.23950)
argues that on long-horizon tasks the harness often affects performance
more than the model does, and that every comparison should disclose it;
kw-bench tests this on long-horizon professional analysis, with the
model held fixed. Code as Agent Harness (arXiv:2605.18747) surveys code
as the operational substrate of agent systems across coding, OS
automation and enterprise workflows. Dive into Claude Code
(arXiv:2604.14228) maps the design space of Claude Code and its
independent peers, each answering the same design questions differently.
Beyond Autonomy (arXiv:2605.10223) takes the governance angle, arguing
for governable, risk-tiered agent execution in the enterprise;
Distributing Security Controls Through Harness Engineering
(arXiv:2607.25890) takes the security angle: the hardened harness, not a
per-team agent configuration, as the unit an organisation builds and
distributes. Our earlier paper (arXiv:2608.20622) argues the same from
enterprise deployment: one harness, unmodified, as the backbone. None of
this work grades a professional deliverable and reusable code on hidden
data, and none holds the model fixed on a long investigative task while
varying the harness.

\textbf{Generative knowledge-work benchmarks.} GDPval (arXiv:2510.04374)
covers 44 occupations with tasks built from the work of industry
professionals, graded by industry professionals against human
deliverables; it also reports that more scaffolding improves
performance. APEX-Agents (arXiv:2601.14242) sets long-horizon,
cross-application tasks from investment banking, consulting and law,
graded with rubrics; the best agent scores 24\%. APEX-Accounting
(arXiv:2607.27189) brings this to accounting --- reconciling accounts,
accruing expenses, posting transactions --- in ten simulated worlds with
rubrics written by accountants; no model exceeds 2.6\% Pass\^{}8.
MBABench (arXiv:2605.22664) grades end-to-end financial spreadsheets on
accuracy, formulas and format. FrontierFinance (arXiv:2604.05912) asks
for financial models that take a skilled human over 18 hours each,
graded by rubric; a second FrontierFinance (arXiv:2608.11683) covers the
investor research workflow with source-attributed rubrics, and finds
that the tool harness, not the model alone, shapes quality and
efficiency. FORCE-Bench (arXiv:2607.19409) grades agentic
enterprise-finance work with rubrics. Finch (arXiv:2512.13168) mines
messy real workflows from the spreadsheets and email of Enron and other
financial institutions, and SpreadsheetBench 2 (arXiv:2606.29955)
evaluates business spreadsheet workflows across large multi-sheet
workbooks. All of them produce something from a brief. In most, the
grader has to decide whether the product is good.

\textbf{Question answering and data analysis.} FinanceBench
(arXiv:2311.11944) and the Finance Agent Benchmark (arXiv:2508.00828)
ask questions about publicly traded companies. DABstep
(arXiv:2506.23719) asks multi-step questions over a financial analytics
platform's data; KramaBench (arXiv:2506.06541) builds pipelines over
data lakes. TaxCalcBench (arXiv:2507.16126) computes individual tax
returns from inputs that are \emph{given} --- our task is finding them.
These grade deterministically, which we share, but on clean data and
short tasks.

\textbf{Agent environments.} TheAgentCompany (arXiv:2412.14161)
simulates a small software company and its consequential everyday tasks.
EnterpriseOps-Gym (arXiv:2603.13594) evaluates stateful agentic planning
and tool use in enterprise settings. Terminal-Bench (arXiv:2601.11868)
is a curated set of hard, realistic tasks in the command line. kw-bench
runs on Harbor, an open framework for evaluating agents in containers.

\textbf{Investigative.} The closest in spirit is Argo-Bench
(arXiv:2610.02122): a simulated food-delivery platform exported to a
235-table ERP warehouse modelled on Oracle E-Business Suite, where the
simulator's ground truth is withheld and the agent must reconstruct
facts before acting. It is investigative, but over one consistently
modelled warehouse. kw-bench adds what real engagements have and clean
warehouses do not: three system generations with partial links between
them, extracts in the formats IT actually delivers, a filed baseline
that is itself wrong, and a deliverable package on company format
templates a specialist would actually sign.

\section*{3. kw-bench: a 1:1 mirror of work that takes
weeks}\label{kw-bench-a-11-mirror-of-work-that-takes-weeks}
\addcontentsline{toc}{section}{3. kw-bench: a 1:1 mirror of work that
takes weeks}

kw-bench runs every harness headless, end to end, with no human in the
loop. The harness is dropped into the problem without any supporting
context from a specialist. It is the ``lazy specialist'' setting: the
controller forwards the email to the harness and walks away. In this
respect the test is unrealistically hard. In real work the specialist
would steer the harness from the outset and throughout, watching what it
does --- ``that's the migration month, check for double posts'', ``the
tool has to take the whole extract folder'', ``this difference is too
big for a quiet month''. The specialist brings judgement; the harness
brings the weeks of evidence work. Our scores are therefore a floor:
what the harness achieves alone. Several of the misses we document in
Section 4 are exactly the ones a specialist would catch in one sentence.

\subsection*{3.1 Where it comes from}\label{where-it-comes-from}
\addcontentsline{toc}{subsection}{3.1 Where it comes from}

We asked tax and forensic specialists for real cases of the kind that
take weeks. Three shaped the design.

The first became the task we ran in this paper. An intra-group chain:
entity A sells to entity B, B does something with the goods, and the
goods ship to an external customer C. Mid-period the client moved from
R/3 to S/4HANA, then hired a consultant to change the process logic
again, leaving three sets of data merged into one, with no reliable
linking tables and no consistent booking logic. Tax had to be recomputed
per jurisdiction, to the cent, over five years. The documentation of how
goods moved did not cover what was needed, so the linking had to be
reverse-engineered. What made the case solvable at all was the returns
filed by the previous adviser --- an anchor to test against.

The second became kw-bench's other task family, procure-to-pay
forensics, which ships in the repository but is not run in this paper.
It is the naive version of a forensic case, and hard nonetheless. A
German subsidiary's cost of materials is about EUR 2.4m over budget, and
the CFO wants a package for Friday's audit committee: a findings
register, a budget-to-actual bridge, correcting journal entries, a
claims register, a memo, and audit analytics that Internal Audit can
rerun at every month-end. About 300 MB of SAP extracts hide four
schemes, each visible only by joining several tables: a ghost vendor
created and paid by one clerk; a bank account diverted for a single
payment run and switched back the next day, traceable only through
change documents and the payment run; price creep by one buyer against
purchase-order prices; and duplicate payments, some partly offset by
credit notes. Next to them sit the honest look-alikes --- legitimate
bank changes, quick reverts, duplicates caught and reversed, a vendor
with a similar name --- and accusing any of them costs points. The
design is the same: a hidden truth, traps, company templates, a tool
graded on data it has never seen, deterministic grading.

The third we deliberately did not build: a full booking dump and the
question why certain accounts had turned positive, with no number and no
process to test any explanation against. The specialists' verdict: the
first answer a model proposes is as valid as the fifteenth. This case is
our argument for why grading needs a hidden, checkable truth. A task
without an anchor cannot tell analysis from storytelling.

\subsection*{3.2 The task: VAT chain
provenance}\label{the-task-vat-chain-provenance}
\addcontentsline{toc}{subsection}{3.2 The task: VAT chain provenance}

Everything in the task is fabricated: the company, its people, its
customers, every document and every posting. Nothing comes from a
client. But it is built to be indistinguishable in shape from the real
cases specialists described to us --- the same system history, the same
kinds of extracts, the same errors and the same traps --- so that a
harness that solves it has done the real work, not a puzzle.

\textbf{The company.} The Rotkehl group makes bicycles. DE01 is the
manufacturer in Schweinfurt and ships everything from its German plant.
AT01 in Vienna is a sales company with an e-bike assembly line,
VAT-registered in Austria and Germany. CH01 in Basel is a Swiss sales
company with a Swiss VAT number only. HK01 is a Hong Kong export hub
with a German VAT registration only from 1 January 2023. TW01 in Taiwan
supplies frames. Customers sit in Poland, Italy, the Czech Republic,
Austria, Germany and Switzerland.

\textbf{The business.} A chain transaction: DE01 sells to a group
company, which resells to the external customer, while the goods travel
once, usually straight from Schweinfurt to the customer. EU VAT allows
only one of the two supplies to be the exempt, ``moving'' supply; the
other is taxable in the country where it takes place (Council Directive
2006/112/EC, Art. 36a). Which leg moves depends on who arranges
transport, read from each leg's Incoterm, and on which VAT number the
middle company quotes. Get it wrong and VAT is owed in a different
country, or not at all where it was paid. The task covers five years,
1,432 chains and 3,084 chain billing documents (928 in ECC, 2,156 in
S/4) across eight chain patterns.

\textbf{The trigger.} Monday 19 January 2026, the Head of Group Tax
writes to her tax manager, CFO in copy (Figure 2). The tax office has
announced a VAT special audit of DE01 for 2021--2025, including the
German registrations of AT01 and HK01; Vienna will likely follow. Before
the audit order arrives she wants the exposure and a decision on
correcting voluntarily. She warns that the returns were filed by the
previous adviser from the system's own tax reports, ``so don't take them
as proof of anything'', and that ``we moved from ECC to S/4 in July 2022
and the consultants rebuilt the chain process in 2024, so the documents
are not linked the same way across the five years.'' The CFO adds that
the S/4 team must be able to rerun the classification every month: hand
it over as a tool.

\begin{figure}
\centering
\pandocbounded{\includegraphics[keepaspectratio]{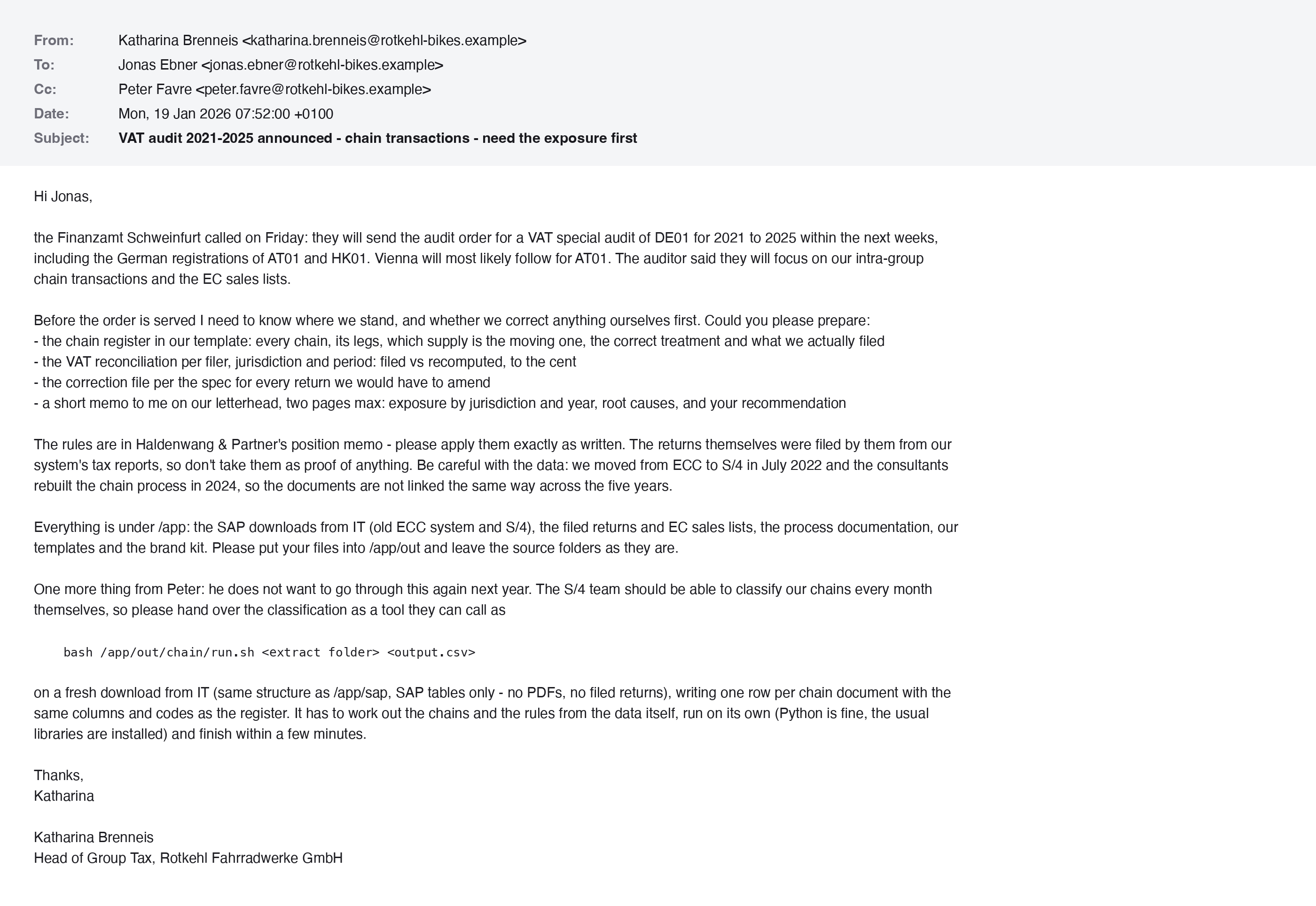}}
\caption{Figure 2. The prompt is an email, exactly as the harness
receives it. It asks for a chain register, a reconciliation to the cent,
a correction file, a two-page memo and a classifier the S/4 team can
rerun.}
\end{figure}

\textbf{The data} --- 62 MB in 83 files in the shape data would arrive
in real world (Figure 3).

\begin{itemize}
\tightlist
\item
  \emph{ECC, 2021 to June 2022.} SE16N text dumps in Windows encoding
  with repeated page headers. Order, delivery and both invoices link
  through the standard document flow.
\item
  \emph{S/4HANA from July 2022.} New number ranges. Open items and open
  deliveries were migrated, but the migration log covers only part of it
  and the old delivery numbers survive on only some deliveries.
  German-decimal CSV, ACDOCA instead of BSEG, MATDOC instead of MSEG.
  Credits for old ECC invoices carry the ECC number in the assignment
  field.
\item
  \emph{The consultant's chain cockpit from 2 April 2024.} The
  intercompany invoice is now billed from a separate chain order, whose
  Incoterm exists only on that order. The two invoices are linked only
  by a custom chain ID --- in a custom field, or in the FI reference
  text --- with hand-typed variants on manually billed documents (about
  14\% of billing is manual). New tax codes K1, K2, K3.
\item
  \emph{The anchor.} The VAT returns and EC sales lists actually filed
  (ELSTER exports with amended submissions, an Austrian FinanzOnline
  CSV), the VAT registrations, and exchange rates --- stored by SAP with
  the date inverted, the monthly VAT rate type as an indirect quotation.
\item
  \emph{The rules.} The previous adviser's position memo states them
  explicitly: moving supply by Incoterm group, the VAT-ID test, the
  billing-date period, the monthly VAT rate with half-up rounding per
  document.
\item
  \emph{Plus} an outdated 2021 process description, the consultant's
  2024 steering-committee deck, the company templates and the brand kit.
\end{itemize}

\begin{figure}
\centering
\pandocbounded{\includegraphics[keepaspectratio]{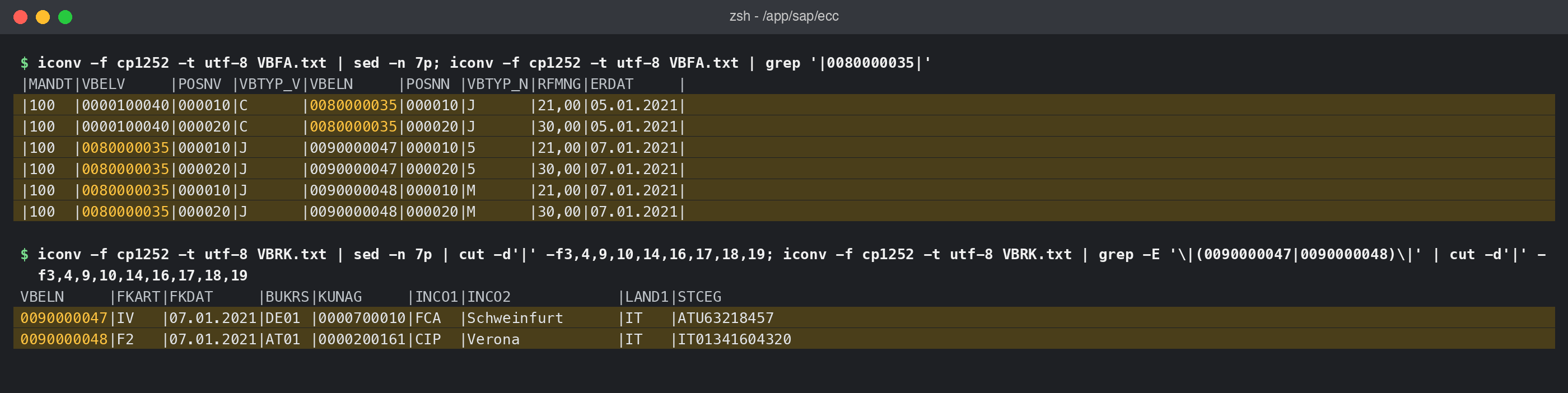}}
\caption{Figure 3a. Regime 1, legacy ECC: one delivery, DE01's
intercompany invoice and AT01's customer invoice, linked through the
document flow. From 2021 to mid-2022 a chain is easy to see.}
\end{figure}

\begin{figure}
\centering
\pandocbounded{\includegraphics[keepaspectratio]{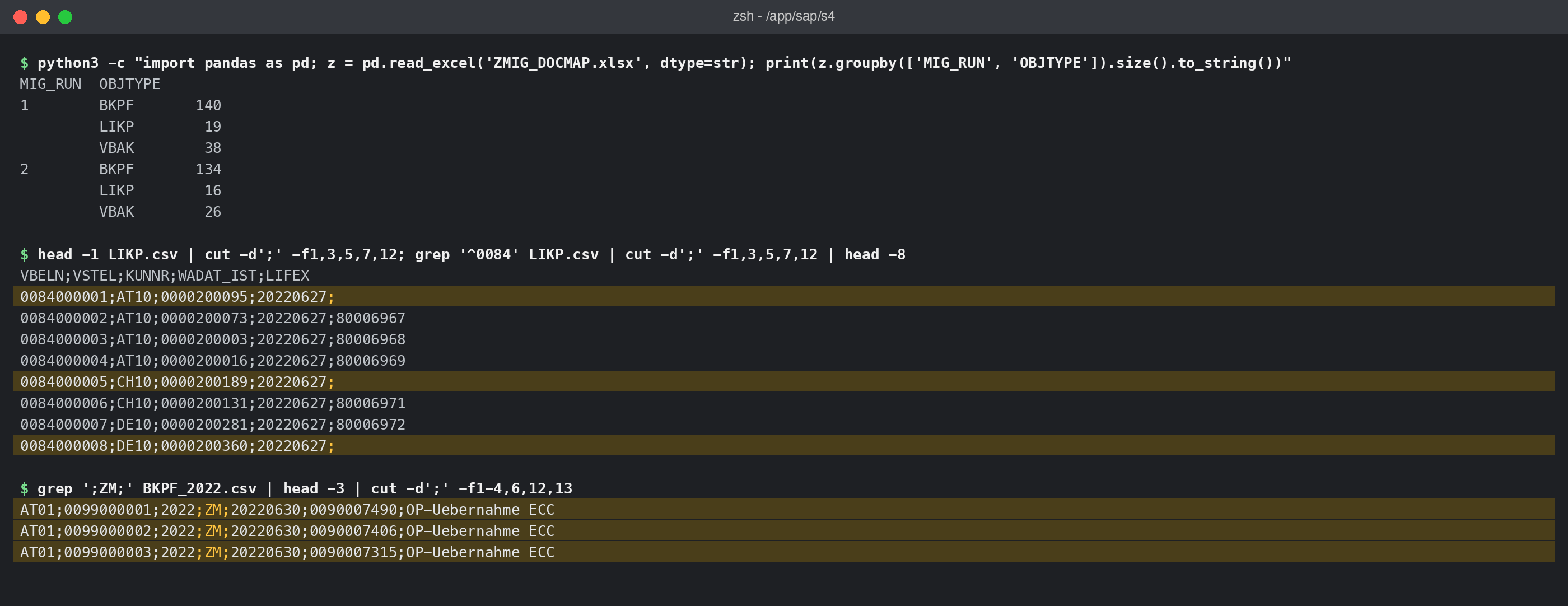}}
\caption{Figure 3b. Regime 2, S/4HANA from July 2022: the migration log
covers only part of the runs, migrated deliveries keep their legacy
number only partly, and open items came over without tax. Linking
mid-2022 documents means combining a partial log with partial legacy
references.}
\end{figure}

\begin{figure}
\centering
\pandocbounded{\includegraphics[keepaspectratio]{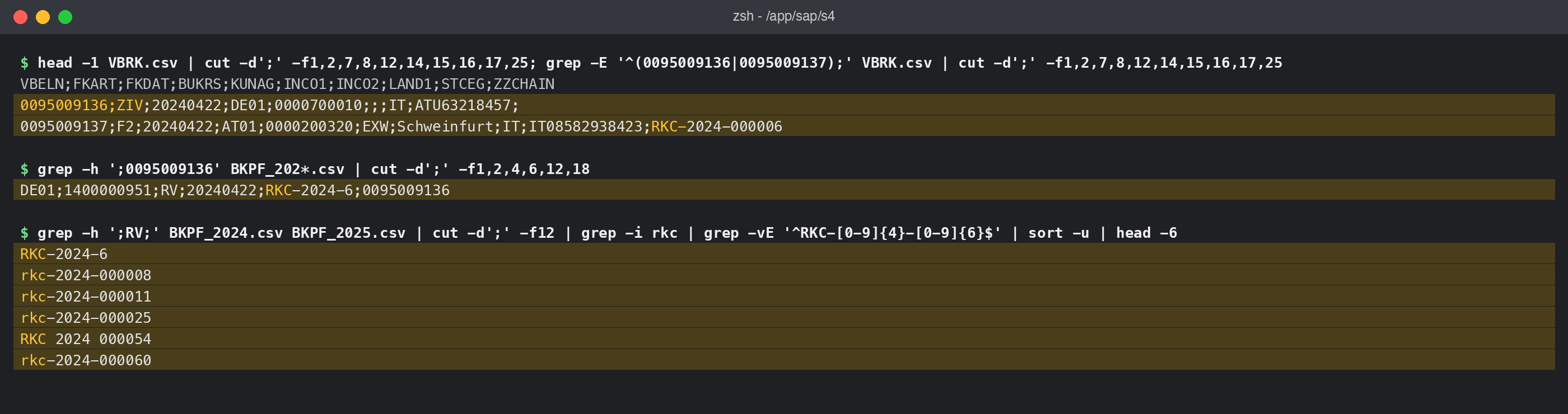}}
\caption{Figure 3c. Regime 3, the 2024 chain cockpit: legs linked only
by a chain ID; the manually billed intercompany leg has no chain ID and
no Incoterm, only a hand-typed reference (``RKC-2024-6'',
``rkc-2024-000008'', ``RKC 2024 000054''). Finding chains now means
normalising what people typed.}
\end{figure}

\textbf{Four planted errors.} The variant we ran plants four mechanisms,
each with a different effect on the filed returns.

\begin{enumerate}
\def\labelenumi{\arabic{enumi}.}
\tightlist
\item
  \emph{Cockpit.} Since 2 April 2024 the cockpit posts every chain leg
  as exempt (K1), whichever supply actually moves, and reports AT01's
  customer-collection chains as triangular (K2). 369 billing documents
  are mis-filed. DE01 under-declared German taxable sales by EUR 14.9m
  and output VAT by EUR 2.83m in 2024--25; AT01's German registration
  under-declared EUR 304,724.83 of output VAT. This is almost the entire
  exposure.
\item
  \emph{Repost.} Eleven June 2022 ECC invoices were migrated as open
  items \emph{and} re-entered by hand in S/4, so they were declared
  twice in July 2022. DE01's July 2022 return is too high: taxable sales
  by EUR 232,680.71, output VAT by EUR 44,209.33, exempt intra-EU
  supplies by EUR 126,001.01. A refund, not a liability, and invisible
  unless the migrated items are matched to their manual re-entries
  (Figure 4a).
\item
  \emph{FX.} S/4 foreign-currency invoices were translated at the daily
  rate instead of the official monthly VAT rate. Dozens of small
  differences across 2022--2025, from cents to about EUR 2,000 per
  return line, a few hundred euros of VAT in total. The test is can
  harness achieve precision to the cent; roughly correct is not enough.
\item
  \emph{Procflag.} The cockpit also gave chain IDs to orders that go
  through AT01's Vienna assembly line, where the goods really move
  twice. AT01 then reported those sales as triangular instead of
  intra-EU supplies and left out its intra-EU acquisitions: EUR 8.8m too
  much triangular, EUR 4.6m too little intra-EU supplies, EUR 4.5m of
  acquisitions missing. No VAT cost, but every EC sales list line is
  wrong (Figure 4b).
\end{enumerate}

\begin{figure}
\centering
\pandocbounded{\includegraphics[keepaspectratio]{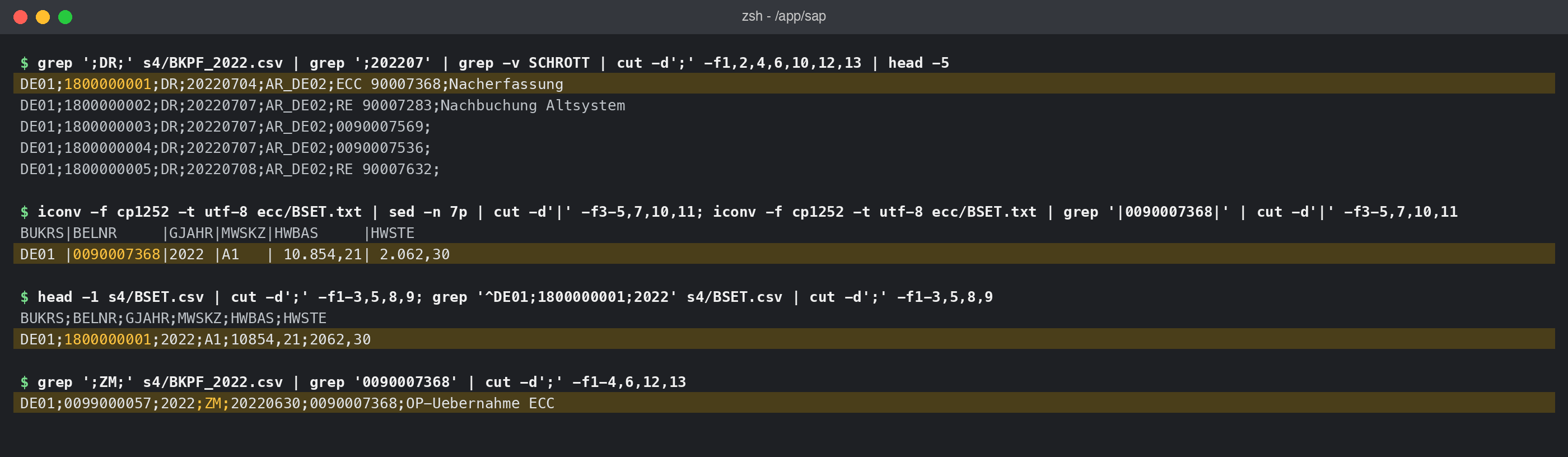}}
\caption{Figure 4a. Evidence trail for the repost error: a June 2022 ECC
invoice re-entered in S/4 (``Nacherfassung'') with the same tax, next to
the migrated open item that carries none. The migration declared it
twice.}
\end{figure}

\begin{figure}
\centering
\pandocbounded{\includegraphics[keepaspectratio]{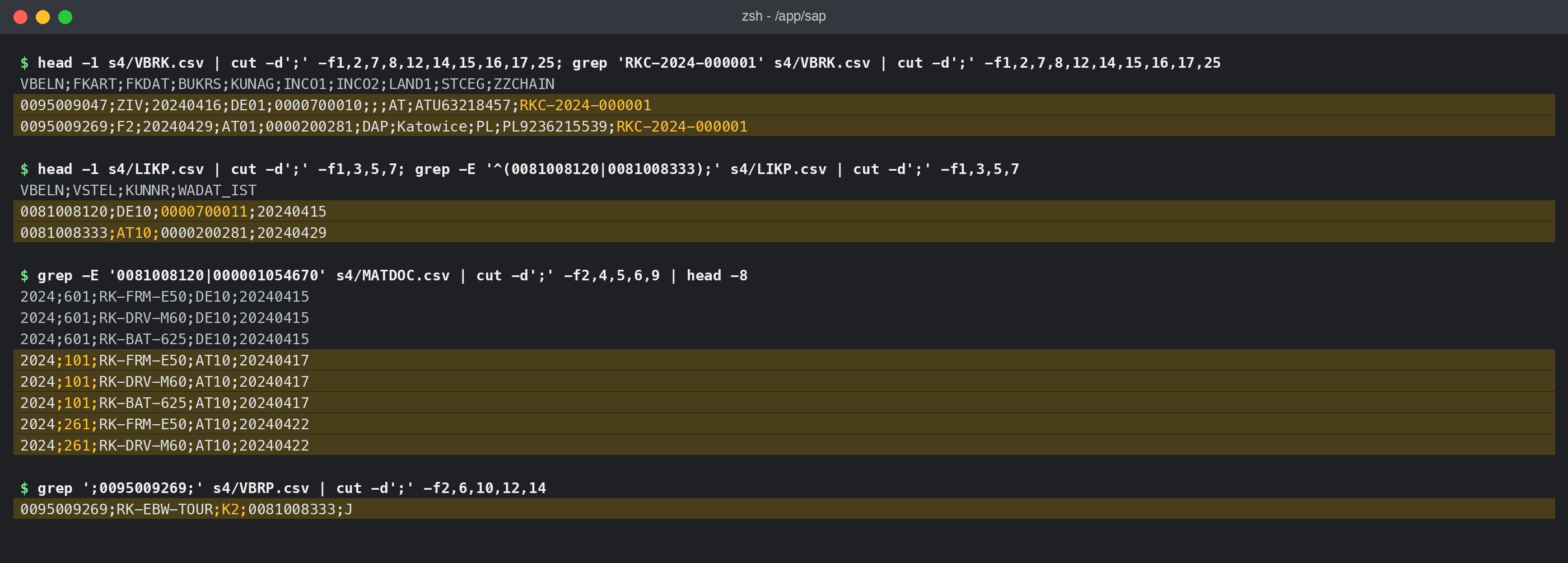}}
\caption{Figure 4b. Evidence trail for the procflag error: a Vienna
assembly order carries a chain ID, but the goods receipt, assembly
consumption and second delivery show two movements. A chain ID is not a
chain.}
\end{figure}

\textbf{The answer.} EUR 3,093,692.38 of additional output VAT: 2022
−44,191.85 (the repost refund and FX), 2023 +176.02 (FX only), 2024
+1,359,579.82, 2025 +1,778,128.39. DE01 EUR 2,788,967.55, AT01's German
registration EUR 304,724.83. 110 returns need amending, all seven lines
each.

\textbf{The traps} --- where naive analysis invents exposure. 717
documents look like chains and are not: 240 stock transfers to Basel and
477 documents of the Vienna assembly flows. Migrated open items carry no
tax. Cancel-and-rebill in the same month. Credit memos across both
systems. AT01 quoting its German VAT number, which legitimately flips
the moving supply. HK01 after its 2023 registration, from when its
exemption is valid. Acquisition tax codes that create two tax lines
each. Late entries that led to amended filings, where the latest filing
counts.

\textbf{The deliverables.} A chain register in the company template with
drop-downs; a VAT reconciliation per filer, jurisdiction, period and
return line, filed against recomputed, to the cent (Figure 5); a
correction upload file per the compliance tool's spec; a memo of at most
two pages on letterhead; and
\texttt{chain/\allowbreak{}run.\allowbreak{}sh}, the monthly classifier,
graded on a hidden dataset with a different seed, different counts and
different registration and go-live dates.

\begin{figure}
\centering
\pandocbounded{\includegraphics[keepaspectratio]{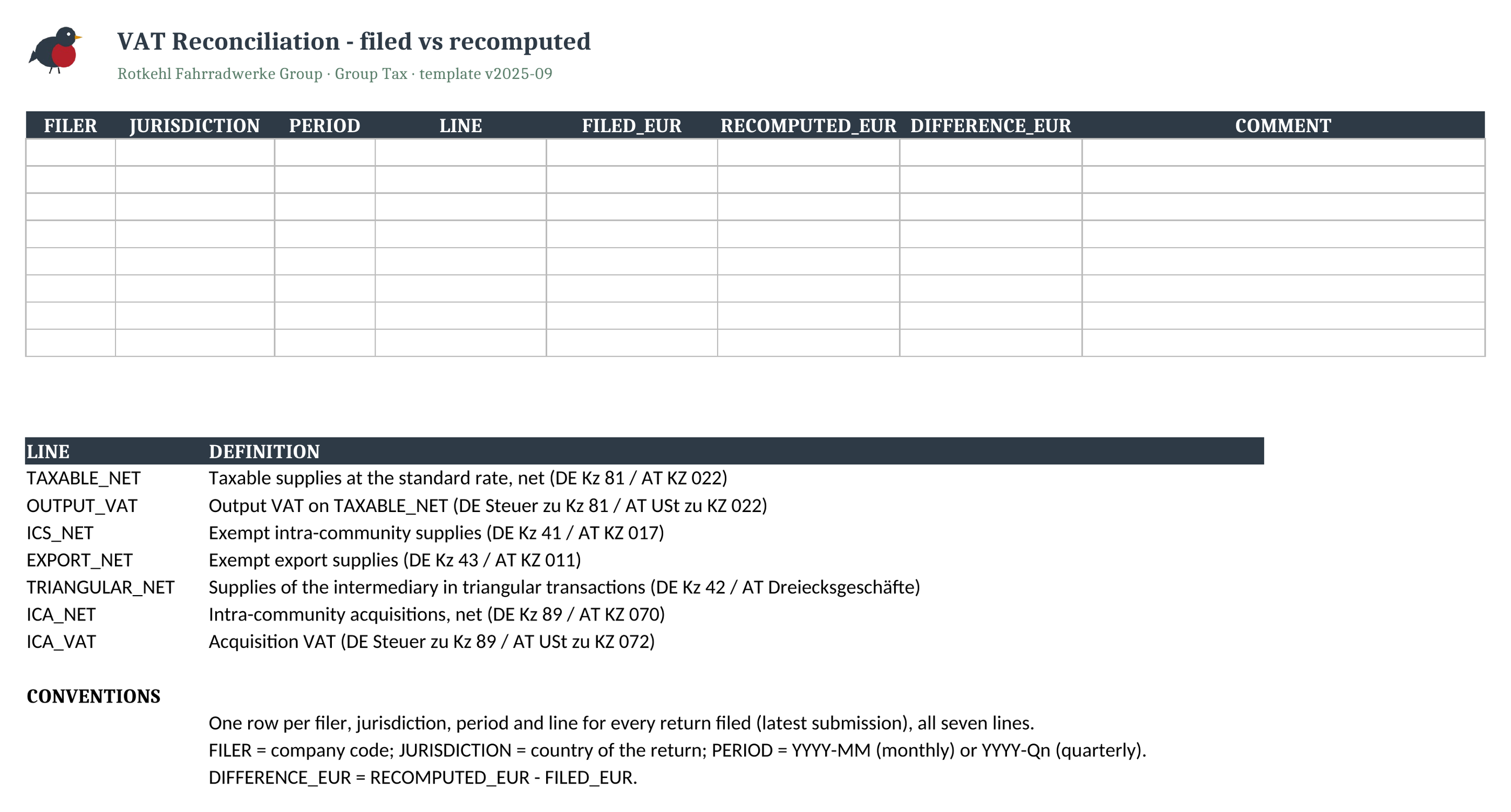}}
\caption{Figure 5. Template for the VAT reconciliation: filed against
recomputed per filer, period and return line. More than a thousand
cells, graded to the cent; the filed returns are the anchor.}
\end{figure}

\subsection*{3.3 The path to the answer}\label{the-path-to-the-answer}
\addcontentsline{toc}{subsection}{3.3 The path to the answer}

Before looking at any harness, we need a yardstick: how would an ideal
team of specialists solve this case in the real world? We wrote it down
as ten steps, each a prerequisite for the next. It is the walk we later
trace every harness against (Section 4.2): where it went, where it
stopped, and why. Our reference solution, which the grader uses to
establish that a perfect score is reachable, follows exactly this path
and scores 1.00. It is about 700 lines of Python and runs in about 30
seconds --- the difficulty is knowing the path, not computing it.

\begin{enumerate}
\def\labelenumi{\arabic{enumi}.}
\tightlist
\item
  \textbf{Read the rules first, not the data.} The adviser's memo fixes
  everything that is a judgement call: transport from each leg's
  Incoterm group, the moving supply from that and from the VAT number
  the middle company quotes, the billing-date period, the monthly rate
  rounded half-up per document. Write it down as a decision table before
  touching a table.
\item
  \textbf{Make the extracts readable.} Strip the page headers, decode
  the Windows encoding, turn ``1.234,56-'' into −1234.56 and inverted
  SAP dates into dates, and load ECC and S/4 into one model that
  remembers which system each row came from.
\item
  \textbf{Tie the filed returns to the books before trusting either.}
  Rebuild what was filed from the posted tax lines (BSET, by tax code)
  and check it equals the submissions, taking the latest amendment per
  period. This proves you can read the returns and gives the baseline
  every difference is measured against.
\item
  \textbf{Reconstruct how documents linked in each era}, and test each
  link rule on its own era: the delivery in ECC; the migration log, then
  the legacy delivery ID, across the cutover; the chain ID in the
  cockpit, from the custom field or the FI reference text, normalising
  hand-typed variants and taking the Incoterm from the chain order; and
  every credit and cancellation back to its original invoice.
\item
  \textbf{Decide what is not a chain.} Two different group companies,
  goods leaving the first seller's own plant, an external final
  customer. That excludes the Basel stock transfers and the Vienna
  assembly flows, where the goods receipt and assembly consumption show
  two movements. Getting this wrong is how most false exposure is
  created.
\item
  \textbf{Classify every chain} with the decision table --- transport,
  moving supply, treatment and jurisdiction of each leg --- applying the
  two VAT-number tests with their dates.
\item
  \textbf{Compare correct with filed treatment, document by document.}
  This is where the cockpit error shows: 369 documents, all from April
  2024, all through the new tax codes.
\item
  \textbf{Recompute each return line from documents}, at the monthly
  rate of the billing month (where FX appears), VAT per document rounded
  half-up. Add what sits outside the chains: acquisitions and manual FI
  customer invoices --- after checking whether a manual invoice is a
  re-entry of a migrated ECC invoice (where the repost appears).
\item
  \textbf{Reconcile and sanity-check.} Recomputed minus filed, per
  filer, jurisdiction, period and line, to the cent. Then look at the
  shape: differences should cluster where a mechanism explains them, and
  quiet months should be quiet. A quarter-million difference in a quiet
  month is a linking bug, not a finding.
\item
  \textbf{Deliver}: register, reconciliation, a correction file for
  exactly the returns that change, a memo that explains each mechanism
  with its amount, and a classifier that takes the same folder structure
  as the extract and was tested on it.
\end{enumerate}

What makes it hard is not any single step but holding all ten
consistent. A linking shortcut in step 4 silently becomes a false
difference in step 9; a trap missed in step 5 becomes invented exposure
in the memo. A human team would split this across a data person, a VAT
specialist and a reviewer, and take weeks. The harness has to be all
three, alone, in two hours.

\subsection*{3.4 Grading}\label{grading}
\addcontentsline{toc}{subsection}{3.4 Grading}

Grading is deterministic, i.e.~we do not use an LLM judge or a rubric.
The grader script reads only the files in
\texttt{/\allowbreak{}app/\allowbreak{}out}, finds them tolerantly
(template name first, then any matching file name), matches headers by
name and accepts German or English number separators..

\Needspace{12\baselineskip}

\textbf{Table 1. Grading components (VAT chain, variant b).}

\begin{longtable}[]{@{}
  >{\raggedright\arraybackslash}p{(\linewidth - 4\tabcolsep) * \real{0.3168}}
  >{\raggedright\arraybackslash}p{(\linewidth - 4\tabcolsep) * \real{0.1386}}
  >{\raggedright\arraybackslash}p{(\linewidth - 4\tabcolsep) * \real{0.5446}}@{}}
\toprule\noalign{}
\begin{minipage}[b]{\linewidth}\raggedright
Component
\end{minipage} & \begin{minipage}[b]{\linewidth}\raggedright
Weight
\end{minipage} & \begin{minipage}[b]{\linewidth}\raggedright
How it is checked
\end{minipage} \\
\midrule\noalign{}
\endhead
\bottomrule\noalign{}
\endlastfoot
Chain register & 0.20 & Per chain billing document: F1 on (document,
treatment, jurisdiction) 0.30; mis-filed documents found, with false
claims penalised 0.35; moving supply 0.15; filed treatment 0.10; chain
grouping 0.10 \\
Reconciliation & 0.22 & Every filer, period and line; cells that differ
must be recomputed within EUR 0.02 (tax lines half by count, half by
amount); false differences penalised; filed values copied \\
Correction file & 0.12 & Every amended return complete (all seven lines)
and correct; extra returns penalised \\
\texttt{run.\allowbreak{}sh} on hidden data & 0.20 & Run as an
unprivileged user on a holdout with another seed, other counts and other
registration and go-live dates; F1 plus moving supply \\
Memo & 0.08 & Total and yearly exposure, root cause per planted
mechanism, recommendation (text via \texttt{pdftotext}) \\
Traps & 0.06 & No assembly or stock-transfer documents in the register,
no false tax differences; credited only alongside real findings \\
Format and branding & 0.12 & Eight checks: template sheets, headers and
drop-downs, header colour, number formats, correction header, memo PDF
of at most two pages with the logo \\
\end{longtable}

Reference points anchor the scale. The oracle scores 1.00. The right
content in plain files, without templates or PDF, scores 0.88. A
regime-blind solution --- the right rules, but only the classic document
flow, with no chain IDs, migration log, re-entries or VAT rate type ---
scores 0.45. Trusting the filed returns scores 0.27; flagging everything
0.20; an empty answer 0.

\subsection*{3.5 Running it}\label{running-it}
\addcontentsline{toc}{subsection}{3.5 Running it}

kw-bench is a set of Harbor tasks. Each task runs in a fresh container,
limited to 4 CPUs and 6 GB of memory, with only
\texttt{/\allowbreak{}app} visible: the inbox, the SAP extracts, the
filed returns, the documents and the templates. Data is generated
deterministically at image build time by a multi-stage build; the
generator, seed and answers never reach the agent. Ground truth, a
generator copy for the holdout and the holdout seeds enter the container
only at verification. The same seed gives a byte-identical
\texttt{/\allowbreak{}app} across builds. The Harbor oracle scores 1.0
on every task, in about 30 seconds plus 9 seconds of verification
including holdout generation.

The VAT family has three tasks:
\texttt{sap-\allowbreak{}vat-\allowbreak{}chain-\allowbreak{}a} (seed
41207; four other mechanisms, EUR 4.47m, 88 returns),
\texttt{sap-\allowbreak{}vat-\allowbreak{}chain-\allowbreak{}b} (seed
58813; the variant above, EUR 3.09m, 110 returns), and
\texttt{sap-\allowbreak{}vat-\allowbreak{}chain-\allowbreak{}clean}
(seed 29641), a control where everything was filed correctly and
restraint is the right answer. We report only variant b here.

The procure-to-pay family described in Section 3.1 ships as three more
tasks in \texttt{datasets/\allowbreak{}kw-\allowbreak{}sap} (two
variants and a clean control, plus an easier tier); we do not run it
here.

Each run also records wall time, model calls, tool calls, tokens, cost,
and memory: a sampler started by the image entrypoint reads
\texttt{/\allowbreak{}proc} once per second and attributes memory to the
harness itself, to the tools it spawns, and to the container, without
modifying Harbor.

\section*{4. Harness runs against the
benchmark}\label{harness-runs-against-the-benchmark}
\addcontentsline{toc}{section}{4. Harness runs against the benchmark}

\subsection*{4.1 Setup}\label{setup}
\addcontentsline{toc}{subsection}{4.1 Setup}

One model for all four harnesses: DeepSeek v4.1 Flash
(\texttt{deepseek/\allowbreak{}deepseek-\allowbreak{}v4.\allowbreak{}1-\allowbreak{}flash-\allowbreak{}1017})
through OpenRouter, reasoning effort high, each harness configured to
the model's 1M-token window. One run per harness, task
\texttt{sap-\allowbreak{}vat-\allowbreak{}chain-\allowbreak{}b},
headless, two-hour limit, Harbor 0.24.0.

Lab harnesses assume their home model, so each configuration had to be
adjusted where its defaults did.

\Needspace{12\baselineskip}

\textbf{Table 2. Harness configuration.}

\begin{longtable}[]{@{}
  >{\raggedright\arraybackslash}p{(\linewidth - 8\tabcolsep) * \real{0.1485}}
  >{\raggedright\arraybackslash}p{(\linewidth - 8\tabcolsep) * \real{0.0693}}
  >{\raggedright\arraybackslash}p{(\linewidth - 8\tabcolsep) * \real{0.2574}}
  >{\raggedright\arraybackslash}p{(\linewidth - 8\tabcolsep) * \real{0.2673}}
  >{\raggedright\arraybackslash}p{(\linewidth - 8\tabcolsep) * \real{0.2574}}@{}}
\toprule\noalign{}
\begin{minipage}[b]{\linewidth}\raggedright
Harness
\end{minipage} & \begin{minipage}[b]{\linewidth}\raggedright
Version
\end{minipage} & \begin{minipage}[b]{\linewidth}\raggedright
Route
\end{minipage} & \begin{minipage}[b]{\linewidth}\raggedright
How the 1M window was set
\end{minipage} & \begin{minipage}[b]{\linewidth}\raggedright
Other adjustments
\end{minipage} \\
\midrule\noalign{}
\endhead
\bottomrule\noalign{}
\endlastfoot
Claude Code (Anthropic) & 2.1.310 & OpenRouter, Anthropic-compatible
endpoint &
\texttt{CLAUDE\_\allowbreak{}CODE\_\allowbreak{}MAX\_\allowbreak{}CONTEXT\_\allowbreak{}TOKENS=\allowbreak{}1000000}
(it does not know the window of a model it does not recognise) & Effort
high, 32k thinking budget; all helper models redirected to DeepSeek \\
Codex CLI (OpenAI) & 0.164.0 & OpenRouter, Responses API & A model
catalog entry declaring 1M (Codex caps unknown models at 272k) & Custom
provider with remote compaction off (Codex compacts through an
OpenAI-only endpoint); its freeform patch and search tools switched off,
as this route cannot call them; Codex then runs with that catalog
entry's system prompt \\
OpenCode & 1.19.4 & OpenRouter, built-in provider & Read from its model
registry (1,048,576) &
\texttt{-\allowbreak{}-\allowbreak{}variant high}; none \\
micro-cc & 0.2.124 & OpenRouter & Read from its model catalog & None \\
\end{longtable}

For the frontier comparison in Section 5.4 we used Claude Opus 5.5
through the Anthropic API, effort medium, same task, same limit, for
micro-cc and for Claude Code in its home setup.

\textbf{Disclosure.} The author built micro-cc. We used version 0.2.124
throughout. It includes general robustness changes (recovery from a
dropped stream and from an empty turn) and a generic instruction in
headless mode to review the work against the original request before
finishing. Nothing in it is task-specific, and it shipped to all users.
In the spirit of arXiv:2605.23950 we disclose the configuration of every
harness, not only its name.

\subsection*{4.2 The walk}\label{the-walk}
\addcontentsline{toc}{subsection}{4.2 The walk}

We traced each run against the ten steps of Section 3.3. Times are
minutes into the run.

\Needspace{12\baselineskip}

\textbf{Table 3. Each harness along the path to the answer (DeepSeek
v4.1 Flash, two hours).}

\begin{longtable}[]{@{}
  >{\raggedright\arraybackslash}p{(\linewidth - 8\tabcolsep) * \real{0.1485}}
  >{\raggedright\arraybackslash}p{(\linewidth - 8\tabcolsep) * \real{0.2277}}
  >{\raggedright\arraybackslash}p{(\linewidth - 8\tabcolsep) * \real{0.2277}}
  >{\raggedright\arraybackslash}p{(\linewidth - 8\tabcolsep) * \real{0.2277}}
  >{\raggedright\arraybackslash}p{(\linewidth - 8\tabcolsep) * \real{0.1683}}@{}}
\toprule\noalign{}
\begin{minipage}[b]{\linewidth}\raggedright
Step
\end{minipage} & \begin{minipage}[b]{\linewidth}\raggedright
OpenCode
\end{minipage} & \begin{minipage}[b]{\linewidth}\raggedright
micro-cc
\end{minipage} & \begin{minipage}[b]{\linewidth}\raggedright
Codex
\end{minipage} & \begin{minipage}[b]{\linewidth}\raggedright
Claude Code
\end{minipage} \\
\midrule\noalign{}
\endhead
\bottomrule\noalign{}
\endlastfoot
1 Rules first & Done (0.3 min) & Partly: one rule (3.9, goods that never
leave Germany) left out & Done, re-read three times & Read many times,
never applied \\
2 Readable extracts & Done & Done (5 min) & Done (2 min) & Only after 94
min \\
3 Tie out the filed returns & Gap diagnosed, tie-out not built & Missed:
``start from the filed amounts and add the delta'' & Done, late (87--111
min) & Not reached \\
4 Linking per era & Partly: 156 documents never linked & Partly: the
same 156 & ECC and early S/4 only; 0 of 1,072 cockpit documents & Not
reached \\
5 What is not a chain & Partly: 162 assembly documents kept & Partly:
162 kept & Never reached the cockpit era & Not reached \\
6 Classify & 90.1\% right & 90.6\% right & 1,860 documents, precision
0.94 & Not reached \\
7 Correct vs filed & 314 of 369, no false flags & 278 of 369, 99 false
flags & 0 of 369 & Not reached \\
8 Recompute each line & Net lines yes; manual invoices and acquisitions
left out & Missed: filed plus deltas & Partly; repost checked in the
wrong table & Not reached \\
9 Sanity-check & July 2022 signal seen, passed by & July 2022 signal
called ``small''; invented refunds accepted & Not reached & Not
reached \\
10 Deliver & Full package in 74 min; correction file from the register &
Full package; 133 returns amended & One engine file, nothing else &
Nothing \\
\end{longtable}

\textbf{OpenCode} printed the whole position memo first (``I have the
rule set. The tax logic is very specific'', 0.7 min) and implemented it
closely, including the domestic-delivery rule; one misreading labelled
150 Swiss-company legs ``located abroad'' where the memo says export,
which costs score but no euros. At 24.8 minutes it diagnosed the steady
gap between billing totals and DE01's filed returns correctly ---
``That's exactly the 19\% VAT rate! \ldots{} manual FI invoices'' ---
and concluded ``We should use BSET to recompute the returns''. It looked
at a few posted tax lines and moved on without building the tie-out. Its
strength is step 8: it recomputed net amounts from billing documents at
the monthly rate (``BSET used UST monthly rate (1.0808), not KURRF
0.91923'', 19.5 min). That alone reproduces 71 of the 103 small FX cells
to the cent and the July 2022 intra-EU difference of −126,001.01 --- as
a by-product of the method, without ever diagnosing either. At 21.8
minutes its own output read ``ICS\_NET: -126,001.01 total, concentrated
in July 2022''; its reasoning went to the DE01 gap and never came back.
It linked by exact chain ID and delivery, missing the 70 cockpit chains
whose manual leg has a blank ID (140 documents) and 16 documents at the
cutover. Of the 369 mis-filed documents it found 314, and every one it
flagged really was mis-filed. It wrote the full package in 74 minutes
and stopped on its own --- but built the correction file from the
register rather than from its own reconciliation: 47 returns, where its
reconciliation showed differences in 127. The memo names only the
cockpit. Headline: EUR 2,696,360.87 against the true EUR 3,093,692.38
(87\%).

\textbf{micro-cc} read the memo with \texttt{pdftotext} three times in
the first two minutes and encoded the rules with references, but left
out rule 3.9: 74 documents of goods that never left Germany came out as
intra-EU supplies, the first alone creating a false difference of EUR
−31,309.31 in January 2021. Step 3 is where it went wrong. At 47 minutes
it compared document totals with the filed return, saw a gap and
reasoned it away: ``this is expected because the returns were prepared
from the accounting system's tax reports (BSET) \ldots{} start from the
filed return amounts and add the delta from chain misclassifications.''
A tie-out from the posted tax lines would have reproduced the filed
returns exactly. Its reconciliation is therefore the filed figure plus
the differences found in the chain list, not a recomputation --- so FX
cannot appear, the July 2022 double posting is never removed, and every
misclassified 2021--23 document becomes an invented refund. At 77.6
minutes it found the one genuine clue to the double posting, the EC
sales list exactly EUR 126,001.01 below the filed intra-EU line, and
called it ``a small discrepancy''. Its linking missed the same 156
documents as OpenCode. It found 278 of the 369 mis-filed documents, with
99 false flags. It delivered the whole package --- first files at 56
minutes, \texttt{run.\allowbreak{}sh} at 80, memo at 84 --- and spent
the last 35 minutes reviewing, as its headless instruction asks. The
review fixed what a requirement check can fix: the tool takes the
extract folder as the email asked and scores 0.89 on hidden data. It did
not fix what needs domain judgement. A EUR 260,405.01 difference in
March 2023, a quiet month whose true difference is EUR 62.85, survived
the sanity check. More review made it more confident, not more correct,
on the numbers. Its memo names all four mechanisms in prose and reports
EUR 2,069,077.69, netting EUR 409,990.82 of 2021--23 ``overpayments''
that do not exist.

\textbf{Codex} read like a careful senior. It parsed both eras in two
minutes, worked out that S/4 billing and FI documents use separate
number ranges and joined them through the FI reference, and --- alone
among the four --- tied the filed returns to the posted tax lines: ``The
posted VAT control totals match the latest filed returns exactly
\ldots{} use BSET for what was filed, then recompute each economic
supply'' (105 min). We verified it independently: to the cent, in every
period. That is the step that makes everything else auditable. But it
never produced a rough package and then improved it. At 46 minutes a
single model call generated 71,447 output tokens of repeated
\texttt{</\allowbreak{}tool>} text, costing 14 minutes; afterwards Codex
started its orientation again, re-reading the memo at 60.6 and 62.6
minutes. Its last reasoning, at 119.9 minutes, says ``the request
involves complex tax rules from a memo I don't have'' --- after reading
the memo three times. At two hours it was still ``rebuilding the engine
properly''. It left one 30 KB
\texttt{chain\_\allowbreak{}engine.\allowbreak{}py}: a working
classifier for ECC and pre-cockpit S/4 (1,860 documents, precision 0.94,
moving supply right on all), none of the 1,072 cockpit documents linked,
so none of the 369 mis-filed found. No register, reconciliation,
correction file, memo or \texttt{run.\allowbreak{}sh}. Score 0.00.

\textbf{Claude Code} never left orientation. For the first 94 minutes it
issued 408 commands inventorying templates, README files, PDFs and
headers; its progress messages almost all restate the same step (``Now
let me inspect the templates and all data file structures''). Its Read
tool returns PDFs as page images, which this model on this route could
not use (``The PDF text didn't render'', 28.4 min; ``The Read tool gave
page counts but not text'', 28.9 min). It worked around that through the
shell --- and kept starting its inventory over. At 94.5 minutes its
context of 967,258 tokens was compacted to 11,348, which took 8.2
minutes, and it started the inventory again. Real data profiling began
at 99.5 minutes: it noticed that 67 of 586 S/4 intercompany bills have
no chain ID --- exactly the manual billings cockpit linking must catch
--- and that the rate table holds two rate types, the first step to the
FX rule. It drew no conclusion from either. One unusable tool result is
exactly what a harness is supposed to absorb; finding another way is the
whole point of giving a model a shell. The failure is that it never
turned exploration into progress.
\texttt{/\allowbreak{}app/\allowbreak{}out} stayed empty. Score 0.00.

\subsection*{4.3 Three failure modes}\label{three-failure-modes}
\addcontentsline{toc}{subsection}{4.3 Three failure modes}

Three harnesses, three ways to fail with the same model. Claude Code
looped: exploration it never turned into progress. Codex went deep:
analysis it never turned into output. micro-cc and OpenCode shipped,
imprecisely: output checked against the brief but not against domain
sense. Knowledge work needs all three --- find the problem, compute it,
ship it.

\subsection*{4.4 Inertia}\label{inertia}
\addcontentsline{toc}{subsection}{4.4 Inertia}

The hard part is not knowing. It is still knowing two hours later.

The walks show that no single step defeated any harness. Every one of
the ten steps was reached, and mostly solved, by at least one of them.
What defeated them was keeping many weak signals alive at once across a
long session: the memo's monthly-rate rule, the July 2022 cutover month,
the hand-typed chain IDs, that 2023 should be a quiet year, the Vienna
assembly exception, the folder structure the tool must accept. Each
harness picked up most of these at some point and then dropped them
while busy with something else. Both delivering harnesses saw the July
2022 double posting and walked past it --- OpenCode at 22 minutes
(``concentrated in July 2022''), micro-cc at 78 (``a small
discrepancy''). micro-cc named FX in its memo and never computed it.
OpenCode diagnosed the missing manual invoices and never added them.
Codex tied the posted tax lines to the filed returns to the cent and
never used that baseline. Claude Code found the blank chain IDs and the
two rate types and never turned either into a rule. This is the
``ignored constraints'' failure of arXiv:2604.13107, stretched over two
hours.

A specialist holds these signals as a checklist and reflects on all of
them before signing. That is where the specialist's role moves: from
doing the evidence work to holding the checklist and steering. Each of
these misses costs a specialist watching the session one sentence.

\section*{5. Results}\label{results}
\addcontentsline{toc}{section}{5. Results}

\subsection*{5.1 Scores}\label{scores}
\addcontentsline{toc}{subsection}{5.1 Scores}

\Needspace{12\baselineskip}

\textbf{Table 4. Score components, DeepSeek v4.1 Flash, two hours
(weights in parentheses).}

\begin{longtable}[]{@{}
  >{\raggedright\arraybackslash}p{(\linewidth - 16\tabcolsep) * \real{0.1313}}
  >{\raggedright\arraybackslash}p{(\linewidth - 16\tabcolsep) * \real{0.1111}}
  >{\raggedright\arraybackslash}p{(\linewidth - 16\tabcolsep) * \real{0.1212}}
  >{\raggedright\arraybackslash}p{(\linewidth - 16\tabcolsep) * \real{0.1010}}
  >{\raggedright\arraybackslash}p{(\linewidth - 16\tabcolsep) * \real{0.1414}}
  >{\raggedright\arraybackslash}p{(\linewidth - 16\tabcolsep) * \real{0.0909}}
  >{\raggedright\arraybackslash}p{(\linewidth - 16\tabcolsep) * \real{0.0909}}
  >{\raggedright\arraybackslash}p{(\linewidth - 16\tabcolsep) * \real{0.1010}}
  >{\raggedright\arraybackslash}p{(\linewidth - 16\tabcolsep) * \real{0.1111}}@{}}
\toprule\noalign{}
\begin{minipage}[b]{\linewidth}\raggedright
Harness
\end{minipage} & \begin{minipage}[b]{\linewidth}\raggedright
Total
\end{minipage} & \begin{minipage}[b]{\linewidth}\raggedright
Register (0.20)
\end{minipage} & \begin{minipage}[b]{\linewidth}\raggedright
Recon (0.22)
\end{minipage} & \begin{minipage}[b]{\linewidth}\raggedright
Correction (0.12)
\end{minipage} & \begin{minipage}[b]{\linewidth}\raggedright
Tool (0.20)
\end{minipage} & \begin{minipage}[b]{\linewidth}\raggedright
Memo (0.08)
\end{minipage} & \begin{minipage}[b]{\linewidth}\raggedright
Traps (0.06)
\end{minipage} & \begin{minipage}[b]{\linewidth}\raggedright
Format (0.12)
\end{minipage} \\
\midrule\noalign{}
\endhead
\bottomrule\noalign{}
\endlastfoot
OpenCode & \textbf{0.542} & 0.81 & 0.30 & 0.00 & 0.88 & 0.24 & 0.00 &
1.00 \\
micro-cc 0.2.124 & \textbf{0.528} & 0.81 & 0.13 & 0.00 & 0.89 & 0.50 &
0.00 & 1.00 \\
Codex & \textbf{0.00} & 0 & 0 & 0 & 0 & 0 & 0 & 0 \\
Claude Code & \textbf{0.00} & 0 & 0 & 0 & 0 & 0 & 0 & 0 \\
\end{longtable}

OpenCode leads micro-cc by 0.014, within what one run per harness can
resolve. Both sit just above the regime-blind reference point (0.45),
and for the same reason: format, tool and register are earned by
classifying documents, and classification was 90\% right. The components
that need exact euro figures --- reconciliation, correction file, traps,
together 0.40 --- contributed 0.07 for OpenCode and 0.03 for micro-cc.
The correction file is all or nothing per return, so a return whose
taxable sales are EUR 4,000 off earns nothing even if six of seven lines
are exact. The grader reads only
\texttt{/\allowbreak{}app/\allowbreak{}out}; analysis that stays in a
transcript, however correct, earns nothing.

\subsection*{5.2 What a specialist gets after two
hours}\label{what-a-specialist-gets-after-two-hours}
\addcontentsline{toc}{subsection}{5.2 What a specialist gets after two
hours}

A score says little about what lands on a specialist's desk. Read as a
work product, the four runs look like this.

\emph{OpenCode} hands over a triage list, a working classifier and a
surprisingly good net-line reconciliation. The register flags 314
mis-filed documents, all of them real: the review list for DE01 and AT01
starts there. 99 of 225 net-line differences are right to the cent,
including every DE01 export cell with an FX difference. The tool scores
F1 0.86 on unseen data. What has to be rebuilt: the tax lines (the
manual invoices are missing), the acquisition lines (copied from the
filed returns, with a comment on every row saying so), and the
correction file, which must come from the reconciliation, not the
register. A specialist sees it from the outputs alone: 47 returns
amended, differences in 127 periods, and an unexplained −126,001.01 in
July 2022.

\emph{micro-cc} hands over a triage list and a working classifier; the
numbers start from scratch. The register gets 91\% of documents right
and flags 278 of the 369 (Figure 6a); the tool scores F1 0.88 on unseen
data and runs in seconds; templates, correction header and memo layout
all pass. But the reconciliation is a delta method, so it cannot be
tweaked, only redone: 156 cells show a difference that does not exist,
the correction file amends 133 returns with none right, and the memo's
EUR 2.07m with its 2021--23 refunds is an artefact (Figure 6b). The tell
is visible without the answer key: ``refunds'' of EUR 109k to 152k in
years where nothing happened, and an Austrian acquisition base eight
times its true size.

\emph{Codex} hands over no deliverable, but not nothing: a fast
classifier for 2021 to early 2024 and, in its transcript, the verified
bridge from the posted tax lines to the filed returns, the FI join and
the acquisition logic. A specialist in a live session could have had a
first draft perhaps an hour or two later. As it stands, nothing can be
handed to anyone.

\emph{Claude Code} hands over an empty folder and a 430-byte helper
script. A few profiling facts in the transcript are correct; they are
inventory, not analysis.

\begin{figure}
\centering
\pandocbounded{\includegraphics[keepaspectratio]{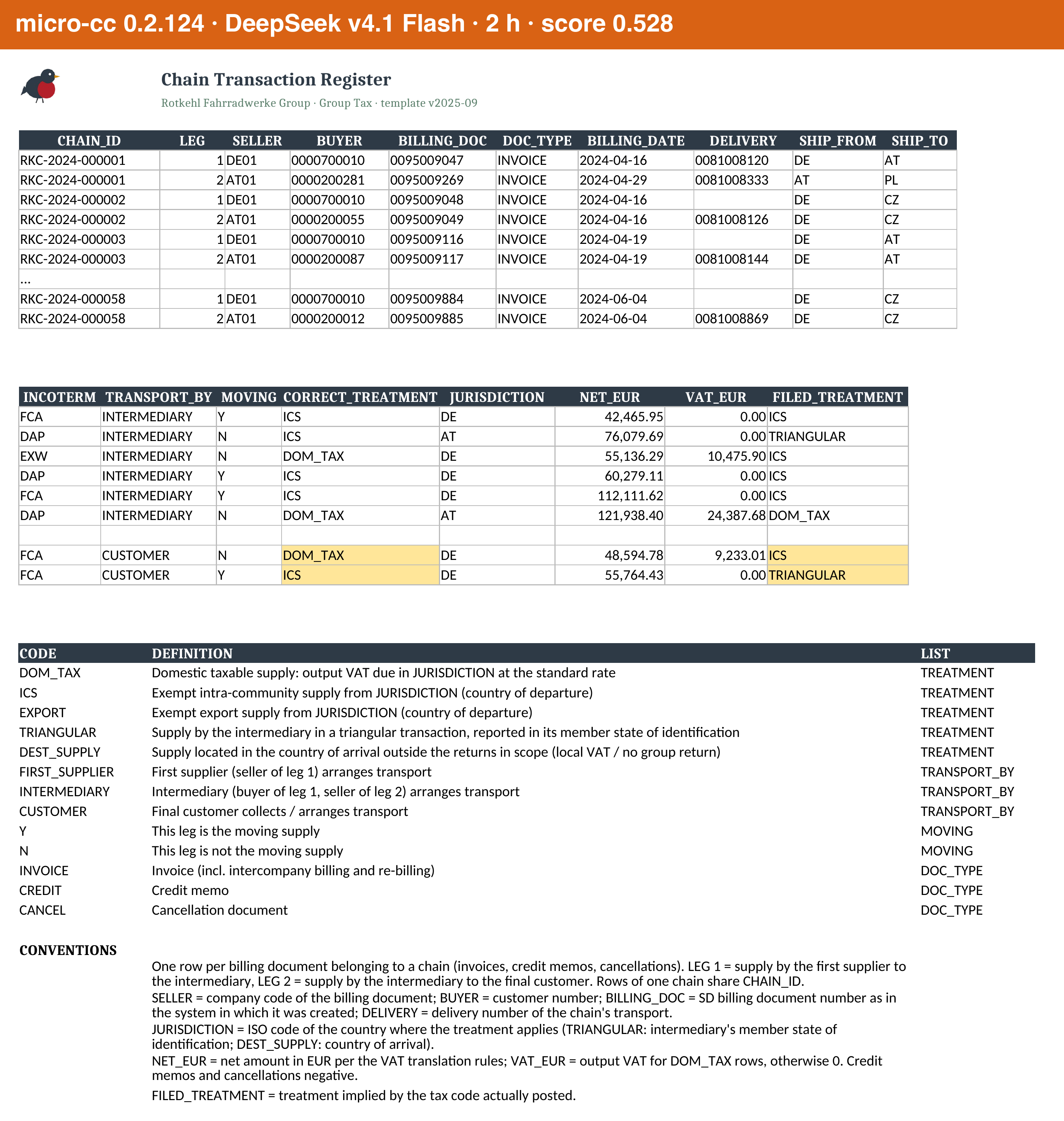}}
\caption{Figure 6a. micro-cc on DeepSeek: the chain register, with a
mis-filed cockpit chain highlighted (filed as intra-EU/triangular,
correct as domestic taxable/intra-EU). The classification layer works:
usable as a triage list.}
\end{figure}

\begin{figure}
\centering
\pandocbounded{\includegraphics[keepaspectratio]{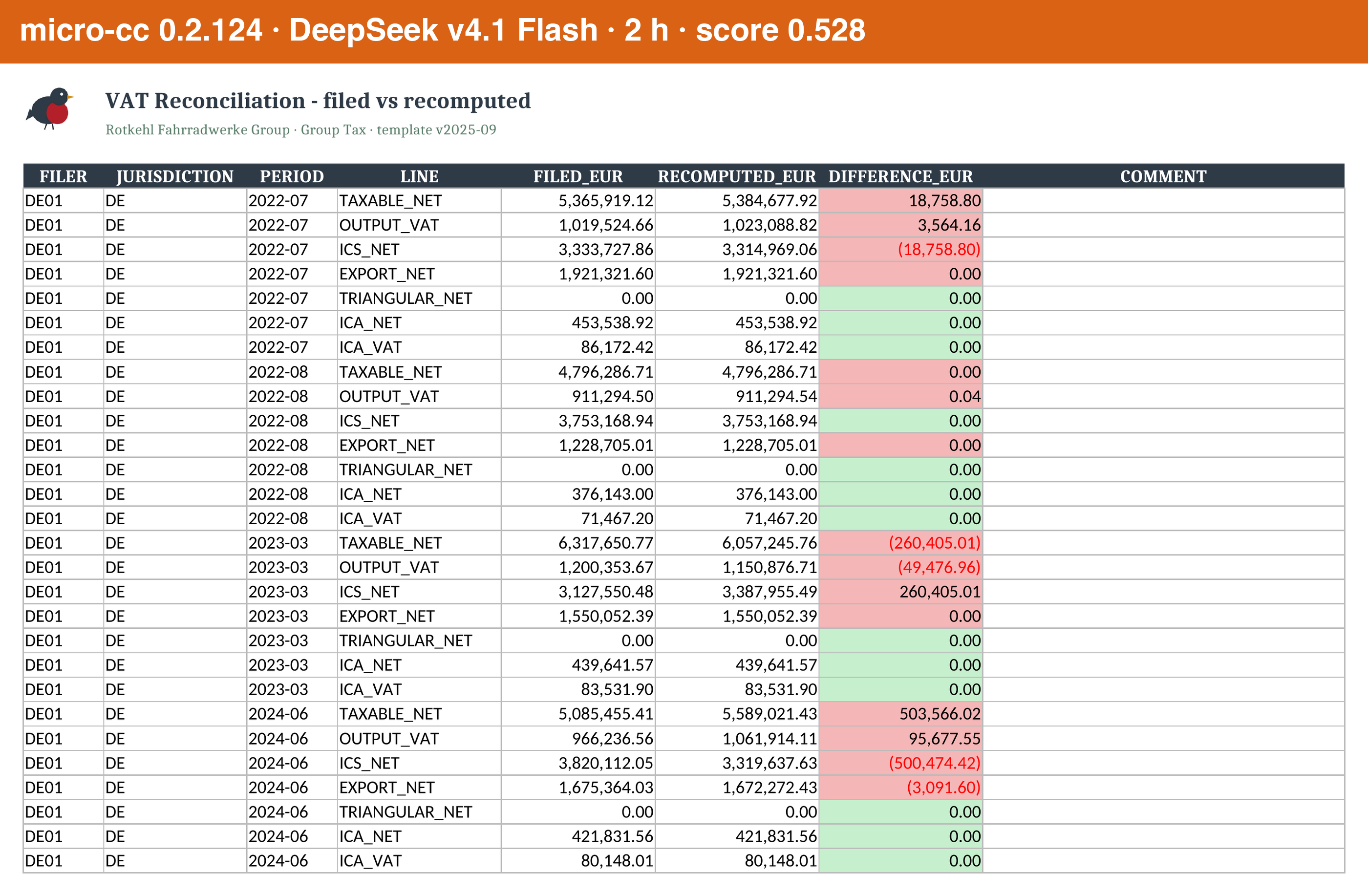}}
\caption{Figure 6b. micro-cc on DeepSeek: the DE01 reconciliation,
differences marked against the answer key (green matches to the cent,
red is wrong). Invented differences in quiet months; the July 2022
double posting never removed.}
\end{figure}

\subsection*{5.3 Why the lab harnesses under-delivered off their home
model}\label{why-the-lab-harnesses-under-delivered-off-their-home-model}
\addcontentsline{toc}{subsection}{5.3 Why the lab harnesses
under-delivered off their home model}

The distinction that matters is lab-built against independent, not
proprietary against open: Codex CLI is open source. The loop is the same
in all four --- a model, a shell, a filesystem. Everything around the
loop is not. From the traces, five differences.

\begin{enumerate}
\def\labelenumi{\arabic{enumi}.}
\tightlist
\item
  \textbf{What goes back into the context.} This is the biggest lever.
  Tool output fed back to the model over the run: micro-cc 277k
  characters, OpenCode 291k, Codex 3.75M, Claude Code 6.78M. Claude
  Code's largest single result was 526,575 characters --- one PDF read
  as page images; the memo, the process note and the correction spec
  were each read several times this way, and every image stays in
  context and is resent on every later call. Codex's largest was a
  725,481-character whole-file dump. Everything returned is re-read on
  every subsequent call: Claude Code averaged about 458k input tokens
  per call over 358 calls, 164M in total. Loose output discipline
  compounds into slowness, cost and confusion.
\item
  \textbf{Tools shaped for the home model.} Claude Code's PDF-as-images
  assumes Claude's vision. Codex's freeform patch and search tools
  assume its own model's tool format; this route could not call them at
  all. With the home model these are advantages; with another,
  liabilities.
\item
  \textbf{Defaults that assume the home model.} Claude Code does not
  know the window of a model it does not recognise; Codex caps unknown
  models at 272k and compacts through an OpenAI-only endpoint. Both
  needed the configuration in Table 2 before they could run at all.
  Without it, every comparison measures the default, not the harness.
\item
  \textbf{Prompts tuned to one model's temperament.} Codex's prompt
  pushes thoroughness; on DeepSeek that became endless depth and no
  output. Claude Code's step-by-step verification became hundreds of
  ``check one more thing'' calls: 178 heredoc Python scripts, nearly all
  reloading the SAP files from scratch, one tool call per turn. A
  correction tuned for one model misfires on another.
\item
  \textbf{Model-agnostic by design.} OpenCode and micro-cc have to work
  on dozens of providers, so their defaults are conservative: plain-text
  tools (both read every PDF through \texttt{pdftotext}), capped tool
  outputs, a context window from a model registry, no vendor-only
  endpoints. They lose nothing on a model that is not a lab's own.
  micro-cc's model wrote a loader module once and imported it; OpenCode
  stopped on its own after 74 minutes and 86 tool calls, for \$0.256.
\end{enumerate}

Lab harnesses are race horses bred for one track. They are co-designed
with their model --- tools, defaults, prompts, context assumptions ---
and taken off that track, the co-design turns into friction. Independent
harnesses are designed for any model, so they lose less. These are
design trade-offs, not bugs: image PDFs, generous outputs and step-wise
verification are the right choices for interactive coding with Claude
and its prompt caching. Unattended, on a long analysis task, with
another model, the same choices become waste. Nor does any of this show
that lab harnesses are worse. It shows that below the frontier, the
harness is the variable that decides.

\subsection*{5.4 So is the task
unsolvable?}\label{so-is-the-task-unsolvable}
\addcontentsline{toc}{subsection}{5.4 So is the task unsolvable?}

Absolutely not.

We ran micro-cc 0.2.124 --- the same version that scored 0.528 --- and
Claude Code, in its home setup, on Claude Opus 5.5 through the Anthropic
API, effort medium, same task, same two-hour limit.

micro-cc scored \textbf{0.9986 in 22 minutes, for \$5.80}. Every chain
document; all 369 mis-filed documents with no false flags; every
reconciliation cell to the cent with no false differences; exactly the
110 returns that need amending, all correct, none extra; F1 0.994 on the
hidden data; no trap documents. Its memo states the exposure as EUR
3,093,692.38 --- the truth, to the cent --- with DE01 at EUR
2,788,967.55, AT01's German registration at EUR 304,724.83, the year
split, all four causes, and the note that the Austrian returns owe no
extra VAT but need their lines corrected. It tied the posted tax lines
to the filed returns within its first twenty or so calls --- step 3,
which no DeepSeek run reached early --- and from then on it was
execution: chain library at 11 minutes, reconciliation at 14,
deliverable builder at 16, memo at 21, done at 22.

Claude Code scored \textbf{1.000 in 24.6 minutes, for \$5.89}. Perfect
on every component, the same EUR 3,093,692.38, identical by year. Its
summary reads like a specialist's: ``The filed returns match SAP's own
tax data to the cent. So every difference below is a wrong tax code or
amount in SAP, not a filing slip.''

On its home model Claude Code is flawless. An independent harness
matches it on the same model, in the same time, for the same money:
0.9986 against 1.000, 22 against 25 minutes, \$5.80 against \$5.89. Even
at home, the context discipline of Section 5.3 shows: Claude Code fed
1.05M characters of tool output back into its context, micro-cc 302k.
Figures 7 and 8 put the same harness side by side with only the model
changed: the delta method becomes a recomputation from the books, and
every red cell turns green.

\begin{figure}
\centering
\pandocbounded{\includegraphics[keepaspectratio]{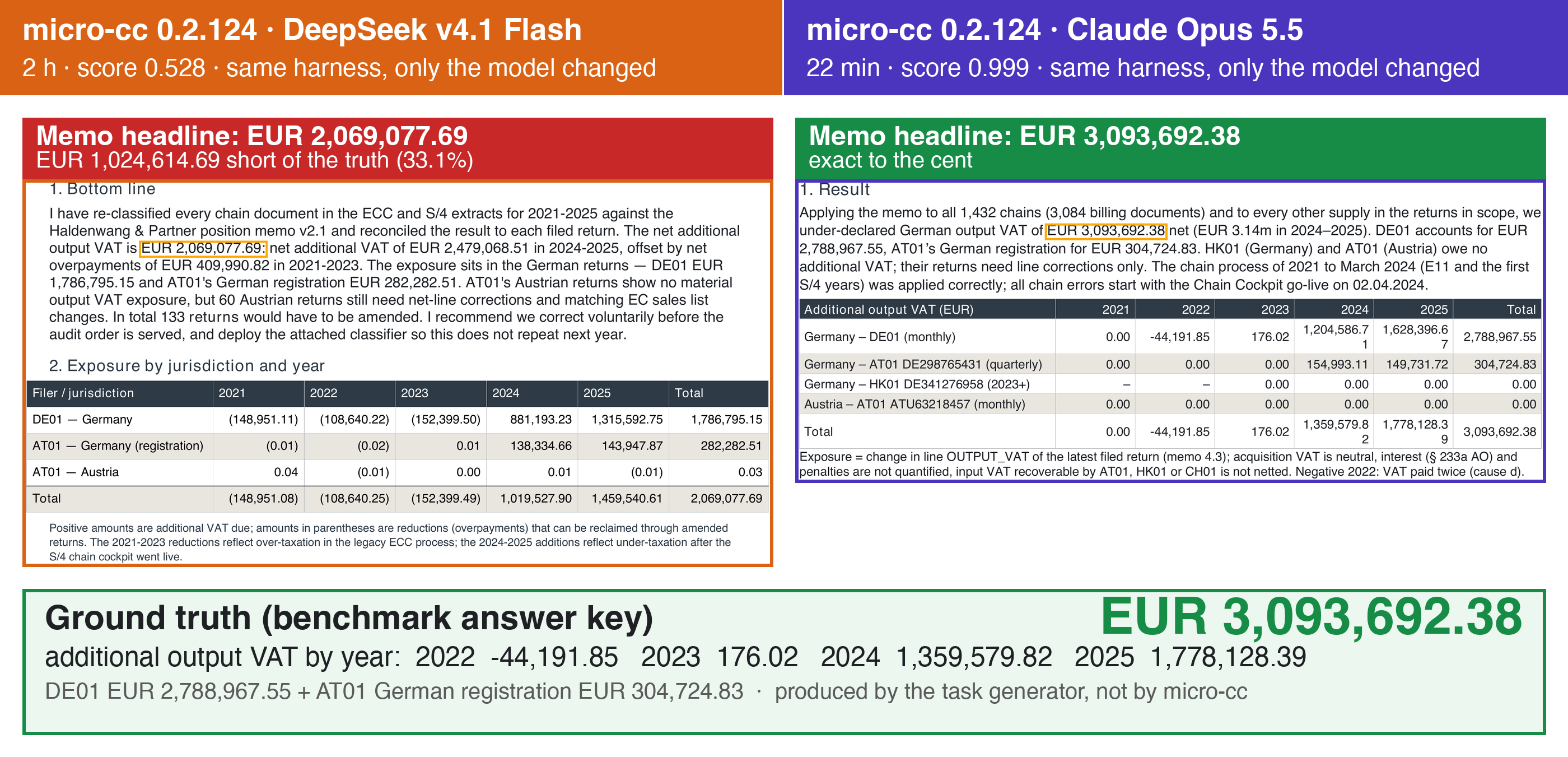}}
\caption{Figure 7. Same harness, only the model changed. micro-cc
0.2.124 on DeepSeek v4.1 Flash (left: EUR 2,069,077.69, 33\% short, with
invented 2021--23 refunds) and on Claude Opus 5.5 (right: EUR
3,093,692.38, exact to the cent). The answer key underneath comes from
the task generator, not from micro-cc.}
\end{figure}

\begin{figure}
\centering
\pandocbounded{\includegraphics[keepaspectratio]{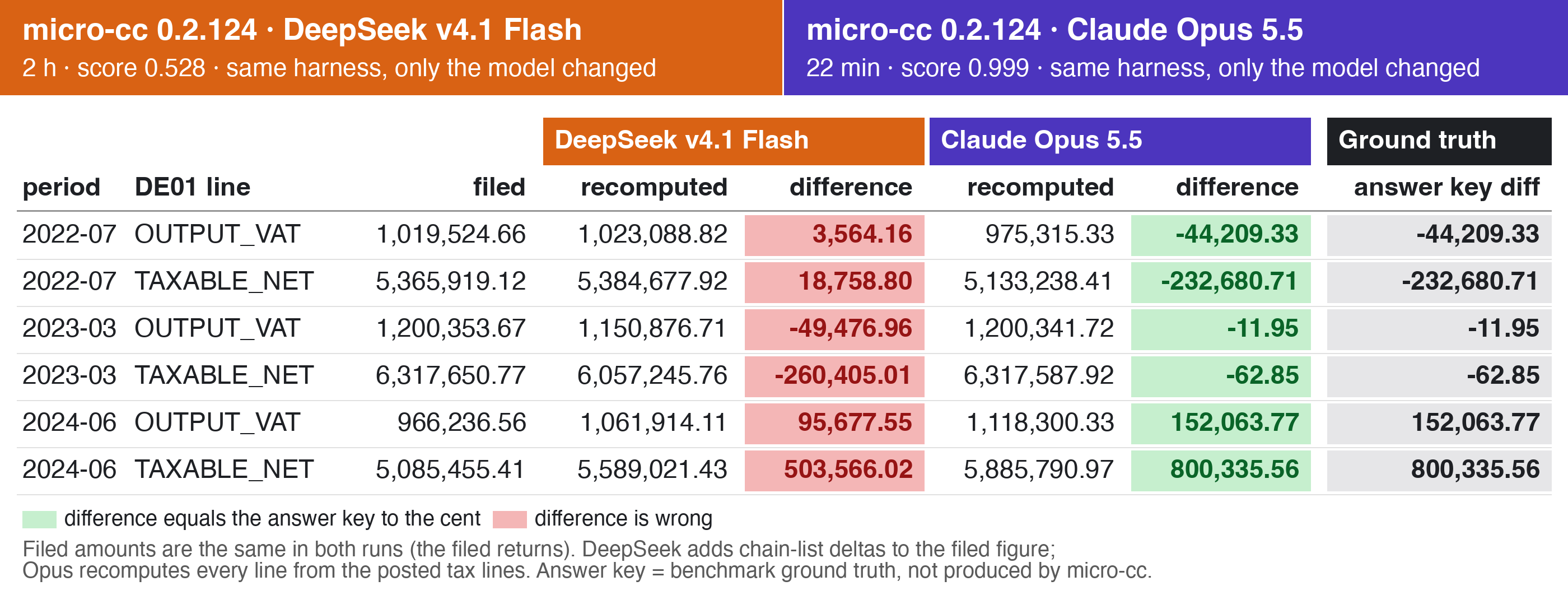}}
\caption{Figure 8. The DE01 reconciliation in both runs against the
answer key. DeepSeek adds chain-list deltas to the filed figure, and
every difference is wrong; Opus recomputes every line from the posted
tax lines, and every difference matches --- including the July 2022
double posting (−44,209.33) and the FX residue of a quiet month
(−11.95).}
\end{figure}

\Needspace{12\baselineskip}

\textbf{Table 5. All six runs.}

\begin{longtable}[]{@{}
  >{\raggedright\arraybackslash}p{(\linewidth - 18\tabcolsep) * \real{0.1300}}
  >{\raggedright\arraybackslash}p{(\linewidth - 18\tabcolsep) * \real{0.0600}}
  >{\raggedright\arraybackslash}p{(\linewidth - 18\tabcolsep) * \real{0.1500}}
  >{\raggedright\arraybackslash}p{(\linewidth - 18\tabcolsep) * \real{0.0700}}
  >{\raggedright\arraybackslash}p{(\linewidth - 18\tabcolsep) * \real{0.0600}}
  >{\raggedright\arraybackslash}p{(\linewidth - 18\tabcolsep) * \real{0.1200}}
  >{\raggedright\arraybackslash}p{(\linewidth - 18\tabcolsep) * \real{0.0800}}
  >{\raggedright\arraybackslash}p{(\linewidth - 18\tabcolsep) * \real{0.1200}}
  >{\raggedright\arraybackslash}p{(\linewidth - 18\tabcolsep) * \real{0.1200}}
  >{\raggedright\arraybackslash}p{(\linewidth - 18\tabcolsep) * \real{0.0900}}@{}}
\toprule\noalign{}
\begin{minipage}[b]{\linewidth}\raggedright
Run
\end{minipage} & \begin{minipage}[b]{\linewidth}\raggedright
Score
\end{minipage} & \begin{minipage}[b]{\linewidth}\raggedright
Wall time
\end{minipage} & \begin{minipage}[b]{\linewidth}\raggedright
Model calls
\end{minipage} & \begin{minipage}[b]{\linewidth}\raggedright
Tool calls
\end{minipage} & \begin{minipage}[b]{\linewidth}\raggedright
Input tokens
\end{minipage} & \begin{minipage}[b]{\linewidth}\raggedright
Output tokens
\end{minipage} & \begin{minipage}[b]{\linewidth}\raggedright
Tool output fed back
\end{minipage} & \begin{minipage}[b]{\linewidth}\raggedright
Cost
\end{minipage} & \begin{minipage}[b]{\linewidth}\raggedright
Harness memory
\end{minipage} \\
\midrule\noalign{}
\endhead
\bottomrule\noalign{}
\endlastfoot
OpenCode · DeepSeek & 0.542 & 74 min (stopped itself) & 83 & 86 & 14.5M
(97\% cached) & 52k & 291k chars & \$0.256 (exact) & 705 MB \\
micro-cc · DeepSeek & 0.528 & 120 min (limit) & 186 & 186 & 45.7M & 502k
& 277k chars & \textasciitilde\$2 (est., \$1.2--2.7) & 147 MB \\
Codex · DeepSeek & 0.00 & 121 min (limit) & 167 & 333 & 65.7M (87\%
cached) & 477k & 3.75M chars & \textasciitilde\$3.3 (est.) & 486 MB \\
Claude Code · DeepSeek & 0.00 & 121 min (limit) & 358 & 445 & 164.0M
(96\% cached) & 180k & 6.78M chars & \textasciitilde\$3.0 (est.) & 343
MB \\
micro-cc · Opus 5.5 & 0.9986 & 22 min (stopped itself) & 74 & 75 &
10.76M & 117k & 302k chars & \$5.80 (exact) & 148 MB \\
Claude Code · Opus 5.5 & 1.000 & 24.6 min (stopped itself) & 79 & 78 &
11.73M & 117k & 1.05M chars & \$5.89 (exact) & 288 MB \\
\end{longtable}

Exact costs come from OpenCode's own per-call accounting, from the
Anthropic usage report (micro-cc on Opus) and from Claude Code's own
accounting, which matches list-price arithmetic. OpenRouter reports
token totals per application but not cost per run, so the other three
DeepSeek costs are estimates: each run's own token counts (uncached
input, cached input, output) priced at the model's list price, then
scaled by the ratio between OpenCode's billed cost and its list-price
cost (1.86). micro-cc does not log its cache share, so its estimate is a
range (85--97\% cached). Harness memory is the harness process alone,
excluding the tools it spawns and the container.

The point is twofold. Below the frontier, the harness decides whether a
cheap model gets halfway or nowhere. At the frontier, a good harness
closes a weeks-long engagement in minutes, for the price of a sandwich.

\section*{6. Discussion and
limitations}\label{discussion-and-limitations}
\addcontentsline{toc}{section}{6. Discussion and limitations}

\textbf{One run per harness.} Every number here is a single run. The
0.014 between OpenCode and micro-cc is within noise; the gap between
delivering and not delivering is not, but its size is not characterised.
Repeated runs and Pass\^{}k are the obvious next step.

\textbf{Two models.} One cheap model and one frontier model. The lab
harnesses were tested off their home model only on DeepSeek; Codex was
not run on an OpenAI model. Whether the gap between independent and lab
harnesses holds on other non-lab models, or narrows as models improve,
is open.

\textbf{Configuration is part of the result.} Lab harnesses needed
configuration changes to run on a model they did not know (Table 2). We
made each one work at the model's real window and disclose every change,
but a different, equally reasonable configuration could change their
results.

\textbf{The author's harness is among those tested.} micro-cc is ours.
Its version includes general robustness changes and a generic review
instruction made during benchmark development (Section 4.1). It did not
win the DeepSeek tier, and on Opus it matched Claude Code; we report it
as one data point among four, and all traces are released.

\textbf{Synthetic data.} The structure comes from real engagements
described by specialists; no client data is used. The tax rules are
simplified and stated in the task: two-leg chains, transport from the
Incoterm, the billing-date period, additional output VAT only, no
interest or penalties. The source cases and the design of the task were
discussed with the tax and forensic specialists who brought them to us.

\textbf{Grading.} The memo is graded by keywords and numbers via
\texttt{pdftotext}, not by a reader. Deterministic grading cannot reward
an insight that is not in a file, which is the point, but it also means
a transcript full of correct analysis (Codex) scores zero.

\textbf{Public ground truth.} The answers for the public tasks ship in
the repository's test folder: hidden from agents during a run, but
public. Contamination of future models is a real risk; the hidden-data
tool test and the ability to regenerate tasks from new seeds are the
mitigation.

\textbf{Headless only.} We measured the lazy specialist. The setting we
expect to matter most --- a specialist steering the harness in a live
session --- is not measured, so the DeepSeek scores are a floor.

\textbf{No human baseline yet.} We state ``weeks'' from the specialists'
account of the real case.

\section*{7. Conclusion}\label{conclusion}
\addcontentsline{toc}{section}{7. Conclusion}

We argued that knowledge workers automate the wrong thing. The slide and
the email are where judgement is expressed and owned; an example of the
expensive part of the work is computing and synthesizing the evidence
underneath across systems --- pulling extracts from systems nobody
documented, linking what three generations of SAP linked differently,
reconciling to the cent. kw-bench mirrors that work one to one, with a
hidden answer the work can be checked against.

Three findings follow. Investigative knowledge work is within reach: on
a frontier model, a forensic VAT engagement that takes a team weeks
collapses into 22 to 25 minutes and six dollars, exact to the cent, with
every deliverable a specialist would sign. Below the frontier, the
harness decides: on the same cheap model the independent harnesses
delivered half the score and the lab harnesses nothing, and the
difference is concrete engineering --- what goes back into the context,
defaults that do not assume a model, tools any model can use. And the
failure mode is inertia, not ignorance: every step was reached by
someone; no one held all of them.

That last finding is also where the specialist comes back in. The
harness does the evidence work. The specialist keeps the judgement ---
holds the checklist, notices the quiet month that is not quiet, says
``that's the migration month, check for double posts'' --- and owns the
conclusion. On the evidence of this benchmark, that division of labour
now costs minutes and dollars.

\section*{Code and supplementary
material}\label{code-and-supplementary-material}
\addcontentsline{toc}{section}{Code and supplementary material}

Code, tasks, grader and reference solution:
\url{https://github.com/GSequist/kw-bench}. Tasks
\texttt{sap-\allowbreak{}vat-\allowbreak{}chain-\allowbreak{}a},
\texttt{sap-\allowbreak{}vat-\allowbreak{}chain-\allowbreak{}b},
\texttt{sap-\allowbreak{}vat-\allowbreak{}chain-\allowbreak{}clean}
(seeds 41207, 58813, 29641) in
\texttt{datasets/\allowbreak{}kw-\allowbreak{}sap-\allowbreak{}vat}; the
procure-to-pay family in
\texttt{datasets/\allowbreak{}kw-\allowbreak{}sap}. A run reproduces
with, for example:

\begin{Shaded}
\begin{Highlighting}[]
\ExtensionTok{harbor}\NormalTok{ run }\AttributeTok{{-}p}\NormalTok{ datasets/kw{-}sap{-}vat }\AttributeTok{{-}a}\NormalTok{ opencode }\AttributeTok{{-}m}\NormalTok{ openrouter/deepseek/deepseek{-}v4.1{-}flash{-}1017 }\AttributeTok{{-}n}\NormalTok{ 1 }\AttributeTok{{-}o}\NormalTok{ jobs}
\end{Highlighting}
\end{Shaded}

micro-cc, the author's harness: \url{https://microcc.xyz},
\url{https://github.com/GSequist/microcc}, and on PyPI as
\texttt{micro-\allowbreak{}cc}.

The Harbor adapter for micro-cc and the configuration files that let
Codex run on a model it does not know (a model catalog entry and a
custom provider) are in the repository's \texttt{agents/\allowbreak{}}
folder. Supplementary material: the run video for micro-cc on DeepSeek
(Figure 1 is its final frame), and the full traces and per-run analyses
of all six runs, available from the author on request.

\section*{References}\label{references}
\addcontentsline{toc}{section}{References}

Bechard, P., Marquez Ayala, O., Chen, E., Skelton, J., Davasam, S.,
Sunkara, S., Yadav, V., Rajeswar, S. Terminal Agents Suffice for
Enterprise Automation. arXiv:2604.00073.
\url{https://arxiv.org/abs/2604.00073}

Benchek, J., Bennett, A., Kern, J., et al.~APEX-Accounting.
arXiv:2607.27189. \url{https://arxiv.org/abs/2607.27189}

Bigeard, A., Nashold, L., Krishnan, R., et al.~Finance Agent Benchmark:
Benchmarking LLMs on Real-world Financial Research Tasks.
arXiv:2508.00828. \url{https://arxiv.org/abs/2508.00828}

Bock, M. R., Molisee, K., Ozer, Z., et al.~TaxCalcBench: Evaluating
Frontier Models on the Tax Calculation Task. arXiv:2507.16126.
\url{https://arxiv.org/abs/2507.16126}

Dong, H., Zhang, P., Gao, Y., et al.~Finch: Benchmarking Finance \&
Accounting across Spreadsheet-Centric Enterprise Workflows.
arXiv:2512.13168. \url{https://arxiv.org/abs/2512.13168}

Egg, A., Iglesias Goyanes, M., Kingma, F., et al.~DABstep: Data Agent
Benchmark for Multi-step Reasoning. arXiv:2506.23719.
\url{https://arxiv.org/abs/2506.23719}

Gore, W. R. Distributing Security Controls Through Harness Engineering.
arXiv:2607.25890. \url{https://arxiv.org/abs/2607.25890}

Islam, P., Kannappan, A., Kiela, D., et al.~FinanceBench: A New
Benchmark for Financial Question Answering. arXiv:2311.11944.
\url{https://arxiv.org/abs/2311.11944}

Ivanov, M., Rana, A., Prabhakaran, G. Can Coding Agents Be General
Agents? arXiv:2604.13107. \url{https://arxiv.org/abs/2604.13107}

Krumdick, M., Reddy, V., Chaudhary, S., et al.~FrontierFinance: A
Long-Horizon Computer-Use Benchmark of Real-World Financial Tasks.
arXiv:2604.05912. \url{https://arxiv.org/abs/2604.05912}

Lai, E., Vitagliano, G., Zhang, Z., et al.~KramaBench: A Benchmark for
AI Systems on Data-to-Insight Pipelines over Data Lakes.
arXiv:2506.06541. \url{https://arxiv.org/abs/2506.06541}

Liu, J., Zhao, X., Shang, X., Shen, Z. Dive into Claude Code: The Design
Space of Today's and Future AI Agent Systems. arXiv:2604.14228.
\url{https://arxiv.org/abs/2604.14228}

Malay, S. K. R., Nayak, S., Nair, J. S., et al.~EnterpriseOps-Gym:
Environments and Evaluations for Stateful Agentic Planning and Tool Use
in Enterprise Settings. arXiv:2603.13594.
\url{https://arxiv.org/abs/2603.13594}

Merrill, M. A., Shaw, A. G., Carlini, N., et al.~Terminal-Bench:
Benchmarking Agents on Hard, Realistic Tasks in Command Line Interfaces.
arXiv:2601.11868. \url{https://arxiv.org/abs/2601.11868}

Ning, X., Tieu, K., Fu, D., et al.~Code as Agent Harness.
arXiv:2605.18747. \url{https://arxiv.org/abs/2605.18747}

Pan, K., Hou, R. Beyond Autonomy: A Dynamic Tiered AgentRunner Framework
for Governable and Resilient Enterprise AI Execution. arXiv:2605.10223.
\url{https://arxiv.org/abs/2605.10223}

Patwardhan, T., Dias, R., Proehl, E., et al.~GDPval: Evaluating AI Model
Performance on Real-World Economically Valuable Tasks. arXiv:2510.04374.
\url{https://arxiv.org/abs/2510.04374}

Pauli, W. M., Panda, S., Admassu, K., et al.~FORCE-Bench: A Benchmark,
Dataset, and Evaluation Harness for Agentic AI in Enterprise Finance.
arXiv:2607.19409. \url{https://arxiv.org/abs/2607.19409}

Salapa, G. J. Applying Anthropic Primitives at Large Enterprises:
Harness Paradigm for Knowledge Work. arXiv:2608.20622.
\url{https://arxiv.org/abs/2608.20622}

Tomitsuka, G., Raayatsanati, A., Xing, E., et al.~Argo-Bench: Evaluating
Data Agents on Enterprise-Scale Workflows. arXiv:2610.02122.
\url{https://arxiv.org/abs/2610.02122}

Vidgen, B., Mann, A., Fennelly, A., et al.~APEX-Agents.
arXiv:2601.14242. \url{https://arxiv.org/abs/2601.14242}

Xu, F. F., Song, Y., Li, B., et al.~TheAgentCompany: Benchmarking LLM
Agents on Consequential Real World Tasks. arXiv:2412.14161.
\url{https://arxiv.org/abs/2412.14161}

Yen, T., Poeltl, J., Gear, H. S., et al.~MBABench: Evaluating LLM Agents
on End-to-End Spreadsheet Tasks in Finance. arXiv:2605.22664.
\url{https://arxiv.org/abs/2605.22664}

Zhang, Y., Koyluoglu, O. O., Venkatesh, T., et al.~FrontierFinance: A
Challenging Benchmark for Measuring Frontier Intelligence of Finance
Agents. arXiv:2608.11683. \url{https://arxiv.org/abs/2608.11683}

Zhang, Y., Wang, J., Ge, Y., Xu, W., Hamm, J., Reddy, C. K. Stop
Comparing LLM Agents Without Disclosing the Harness. arXiv:2605.23950.
\url{https://arxiv.org/abs/2605.23950}

Zhu, J., Zhang, Y., Ma, Z., et al.~SpreadsheetBench 2: Evaluating Agents
on End-to-End Business Spreadsheet Workflows. arXiv:2606.29955.
\url{https://arxiv.org/abs/2606.29955}

Council of the European Union. Council Directive 2006/112/EC of 28
November 2006 on the common system of value added tax, Art. 36a (chain
transactions), as inserted by Council Directive (EU) 2018/1910.
\url{https://eur-lex.europa.eu/eli/dir/2006/112/oj}

Harbor. Framework for evaluating and improving agents.
\url{https://github.com/harbor-framework/harbor}

Software. Claude Code, Anthropic:
\url{https://github.com/anthropics/claude-code}. Codex CLI, OpenAI
(Apache-2.0): \url{https://github.com/openai/codex}. OpenCode:
\url{https://opencode.ai}. micro-cc: \url{https://microcc.xyz},
\url{https://github.com/GSequist/microcc},
\url{https://pypi.org/project/micro-cc/}.

\end{document}